\documentclass[a4paper,fleqn]{cas-dc}

\usepackage[american]{babel}
\usepackage{csquotes}
\usepackage[backend=biber,style=apa]{biblatex}
\let\cite\parencite
 
\usepackage{placeins} 
\usepackage{adjustbox} 
\usepackage{threeparttable} 
\usepackage{subcaption} 

\def\tsc#1{\csdef{#1}{\textsc{\lowercase{#1}}\xspace}}
\tsc{WGM}
\tsc{QE}
\tsc{EP}
\tsc{PMS}
\tsc{BEC}
\tsc{DE}

\begin{document}
\let\WriteBookmarks\relax
\def\floatpagepagefraction{1}
\def\textpagefraction{.001}
\shorttitle{Semantic Probe for SF-CDFSL} 
\shortauthors{Yaze Zhao et~al.}

\title [mode = title]{Reviving In-domain  Fine-tuning  Methods  for Source-Free  Cross-domain Few-shot Learning}                      



\author{Yaze Zhao}[style=chinese, 
                        orcid=0009-0000-8810-8929, 
                        ] 
\fnmark[1]   
\ead{zyaz@hust.edu.cn}
\credit{Writing – review and editing, Writing – original draft,
Visualization, Validation, Methodology, Investigation, Formal analysis,
Data curation, Conceptualization}


\affiliation{organization={School of Computer Science and Technology, Huazhong University of Science and Technology},
                addressline={1037 Luoyu Road}, 
                city={Wuhan},
                postcode={430070}, 
                state={Hubei},
                country={China}}

\author{Yicong Liu}[style=chinese, 
                        orcid=0009-0003-9402-503X, 
]
\ead{lyc.hust@foxmail.com}
\fnmark[1]   
\credit{Writing – review and editing, Writing – original draft,
Visualization, Validation, Methodology, Investigation, Formal analysis,
Data curation, Conceptualization}

\author{Yixiong Zou}[%
   style=chinese,
 orcid=0000-0002-2125-9041
   ]
\cormark[1]   
\ead{yixiongz@hust.edu.cn}
\credit{Writing – review and editing, Supervision, Methodology, Funding acquisition}

\author{Yuhua Li}[style=chinese,
 orcid=0000-0002-1846-4941
]
\ead{idcliyuhua@hust.edu.cn}
\credit{Resources, Project administration}

\author{Ruixuan Li}[style=chinese, 
 orcid=0000-0002-7791-5511
]
\ead{rxli@hust.edu.cn}
\credit{Resources, Project administration}

\fntext[fn1]{These authors contributed equally to this work.}
\cortext[cor1]{Corresponding author}



\begin{abstract}
Cross-Domain Few-Shot Learning (CDFSL) aims to adapt large-scale pretrained models to specialized target domains with limited samples, yet the few-shot fine-tuning of vision-language models like CLIP remains underexplored. By establishing multiple fine-tuning baselines of CLIP for CDFSL, we find adapter-based methods (e.g., LoRA) consistently outperform prompt-based ones (e.g., MaPLe) among contemporaneous representative methods—contrary to in-domain scenarios. To make those effective in-domain methods competitive  again in CDFSL, we analyze this phenomenon and discover LoRA’s superiority stems from rectifying the collapsed attention of visual [CLS] token, enhancing modality alignment and class separation by focusing on text-related visual regions. Further, we find textual [EOS] token exhibits much better attention to visual samples, and CLIP’s standard contrastive loss weakly constrains modality alignment. Based on these insights, we propose Semantic Probe, a plug-and-play attention rectification framework for both adapter- and prompt-based methods. 
Extensive experiments on four CDFSL benchmarks validate our rationale, achieving 
 strong 
 performance and benefiting both fine-tuning paradigms. Codes will be released. 
\end{abstract}

\begin{graphicalabstract}
\begin{center}
\includegraphics[width=0.8\textwidth]{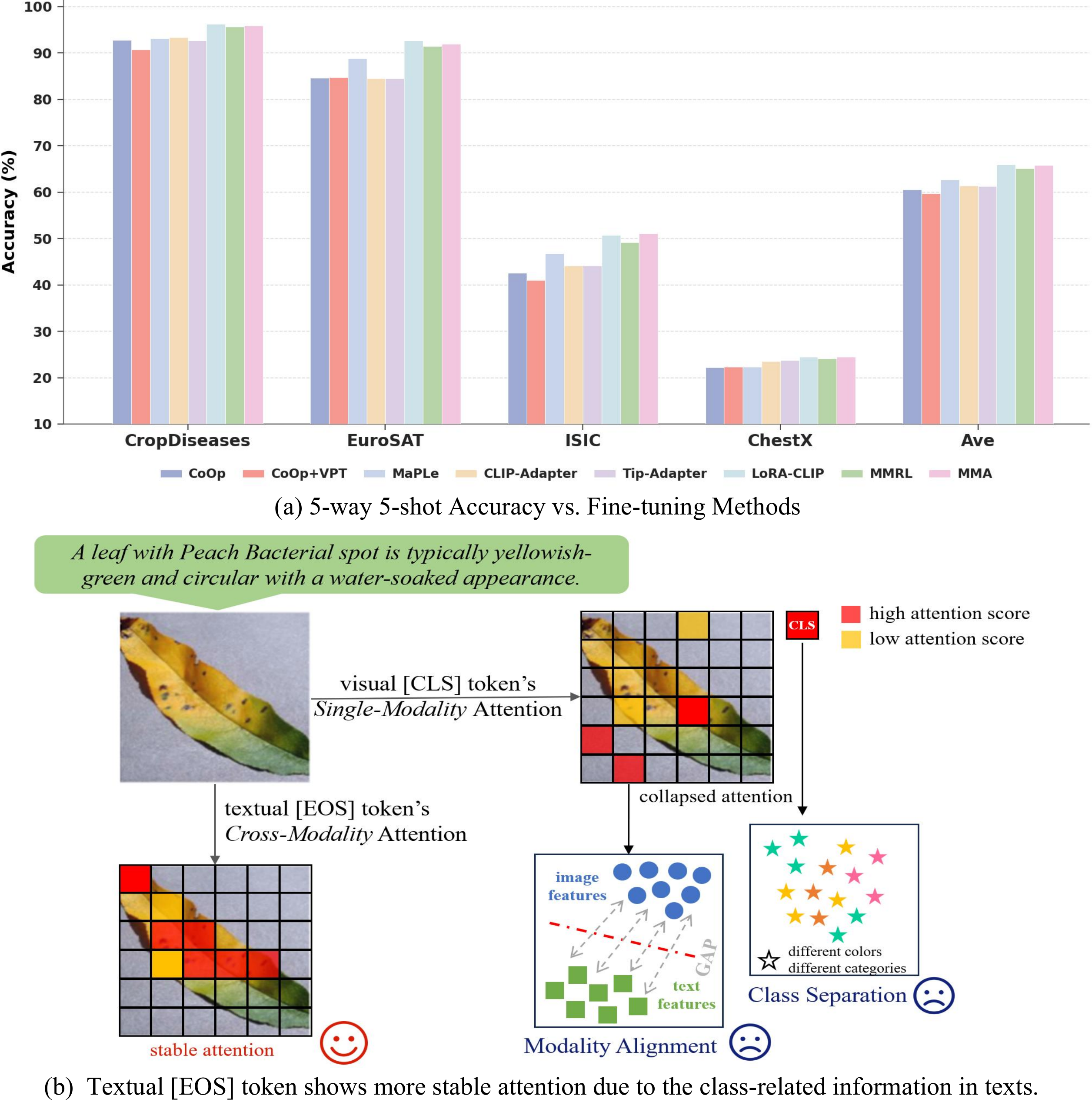}
\end{center}

Cross-Domain Few-Shot Learning (CDFSL) aims to adapt large-scale pretrained models to specialized target domains with limited samples, yet the few-shot fine-tuning of vision-language models like CLIP remains underexplored. By establishing multiple fine-tuning baselines of CLIP for CDFSL, we find adapter-based methods (e.g., LoRA) consistently outperform prompt-based ones (e.g., MaPLe) among contemporaneous representative methods—contrary to in-domain scenarios. To make those effective in-domain methods competitive  again in CDFSL, we analyze this phenomenon and discover LoRA’s superiority stems from rectifying the collapsed attention of visual [CLS] token, enhancing modality alignment and class separation by focusing on text-related visual regions. Further, we find textual [EOS] token exhibits much better attention to visual samples, and CLIP’s standard contrastive loss weakly constrains modality alignment. Based on these insights, we propose Semantic Probe, a plug-and-play attention rectification framework for both adapter- and prompt-based methods. 
Extensive experiments on four CDFSL benchmarks validate our rationale, achieving 
strong  
performance and benefiting both fine-tuning paradigms. 
\end{graphicalabstract}

\begin{highlights}

\item We first establish multiple CLIP fine-tuning baselines for source-free cross-domain few-shot learning (SF-CDFSL) and observe a performance reversal relative to in-domain settings: among contemporaneous representative methods, adapter-based methods (e.g., LoRA) generally outperform prompt-based counterparts (e.g.,  MaPLe).
\item We reveal that the performance reversal stems from the attention collapse of the visual token, which causes severe modality misalignment and poor class separation; meanwhile, the textual token shows more stable and discriminative attention to visual regions. 
\item We propose a plug-and-play Semantic Probe framework consisting of an EOS-guided Attention Rectification (EAR) module and a dynamic Balanced Alignment and Separation (BAS) loss, which revives in-domain fine-tuning methods and achieves 
superior 
performance on four CDFSL benchmarks. 
\end{highlights}


\maketitle 

\section{Introduction}
\begin{figure}
\centering 
\includegraphics[width=0.95\linewidth]{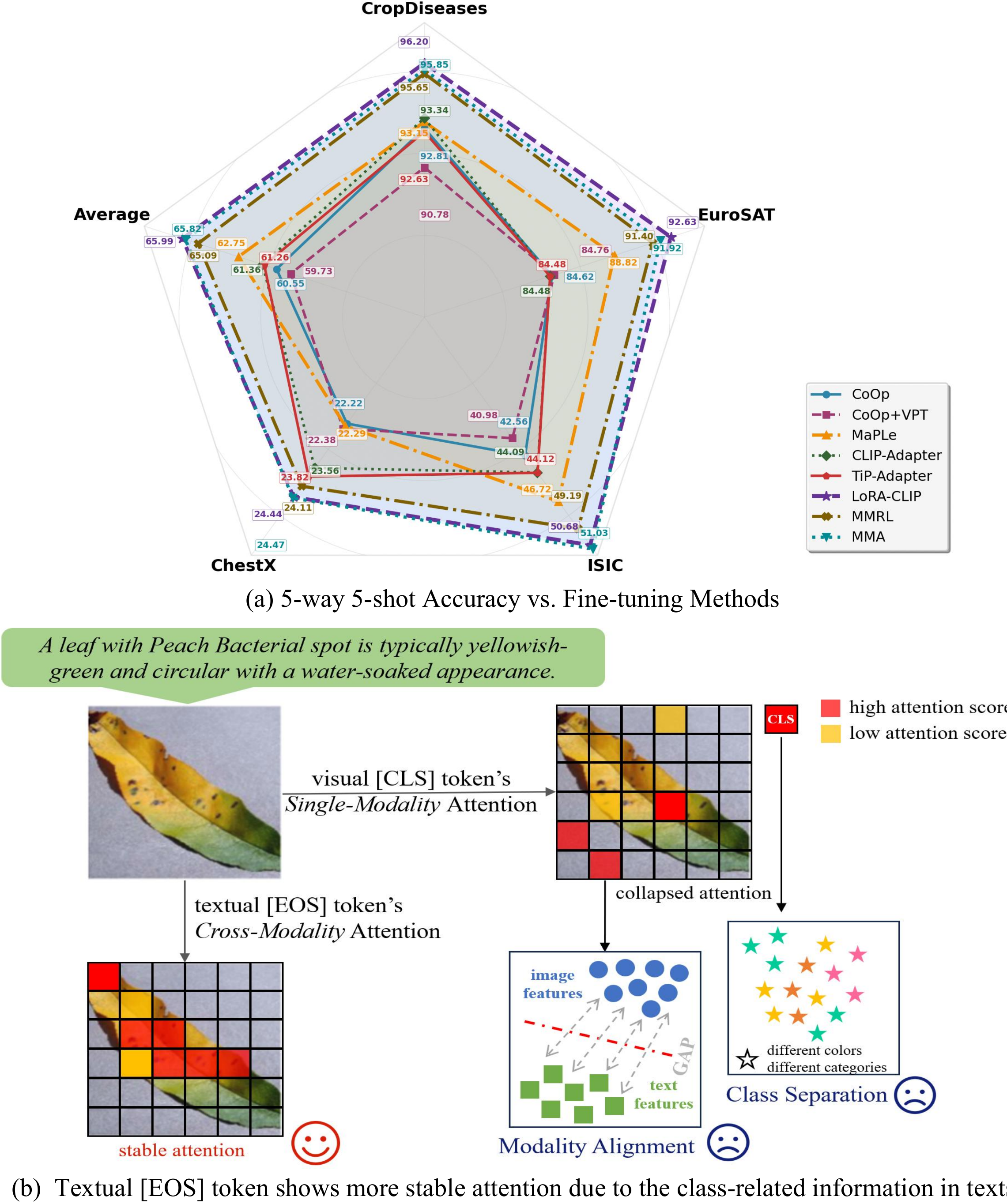}  
\caption 
{
(a) Among these fine-tuning methods, we find LoRA's performance in the CDFSL task is consistently higher than that of others, which is opposite to the in-domain scenarios. 
(b) 
To understand this reversal, we uncover that 
the cause 
 lies in 
the collapsed  visual  attention on cross-domain samples, preventing the model from 
focusing on visual regions relevant to the given class-related text, while LoRA can 
better handle than 
other methods. 
Based on this, we further find that the EOS token of texts  
     inherently maintains 
    more stable attention 
    on discriminative image areas, 
    which inspires us to develop an EOS-guided 
    attention rectification mechanism that can be plugged into both adapter- and prompt-based fine-tuning, making 
    the latter 
    competitive again in CDFSL.  
} 
\label{fig:motivation} 
\vspace{-0.5cm} 
\end{figure}

The past decade has seen significant progress in computer vision~\cite{dosovitskiy2021an, 
 DBLP:conf/cvpr/YangD0D25}. However, this success relies heavily on large-scale, well-annotated datasets like ImageNet, which are widely available in general domains. 
 In contrast, specialized real-world applications, such as 
medical diagnosis and remote sensing, 
 often face severe data scarcity. The high cost of expert annotations, coupled with stringent privacy and security constraints, typically limits labeled samples to only a few per class, creating significant challenges for deploying  robust AI models in these critical domains. Consequently, 
 Cross-Domain Few-Shot Learning (CDFSL)~\cite{guo2020broader, zou2024a, DBLP:journals/corr/abs-2412-00767} has emerged as a critical paradigm to 
 tackle this. 
It aims to transfer models pretrained on large-scale general datasets  
to downstream cross-domain expert datasets  
where only scarce training samples are available. With the increasing adoption of large 
pretrained models, source-free fine-tuning~\cite{DBLP:conf/cvpr/YazdanpanahM22, 
DBLP:journals/tip/XuLZFSCL24, DBLP:journals/corr/abs-2411-10070} of such models has become a 
critical concern for CDFSL. However, vision-language models       
 exemplified by CLIP~\cite{radford2021learning}, which have been extensively applied in various 
 vision tasks due to their remarkable generalizability, remain relatively underexplored for the 
 downstream few-shot fine-tuning  in CDFSL tasks.

To address this issue, 
we first establish multiple downstream few-shot fine-tuning baselines of CLIP by implementing 
different state-of-the-art methods, including
prompt-based tuning \cite{zhou2022learning, jia2022visual, khattak2023maple}, and adapter-based 
tuning \cite{gao2024clip, zhang2021tip, 
zanella2024low} for CLIP, as shown in Fig.\ref{fig:motivation}a. 
However, 
prior studies~\cite{khattak2023self} have shown that prompt-based methods typically outperform adapter-based methods in in-domain tasks. 
In contrast, among representative methods introduced during the same period, our results show 
that adapter-based methods generally outperform prompt-based methods in downstream cross-domain 
scenarios, with LoRA-CLIP achieving the highest average performance. 
This phenomenon indicates that the few-shot fine-tuning for cross-domain scenarios may exhibit 
different mechanisms compared with in-domain ones, which have been seldom studied. 

To delve into this phenomenon, 
in this paper,
we choose LoRA-CLIP   and MaPLe  
as two top-performing methods from  
Fig.~\ref{fig:motivation}a
for an in-depth analysis. 
Given models fine-tuned with these methods, we first measure the feature distribution of text (augmented with LLM-generated~\cite{pratt2023does} prompt expansions)
 and visual modalities, and we find LoRA-CLIP's modality alignment and class separation are much better than those of MaPLe.

To take a closer look at this phenomenon, we then visualize the attention map of visual features. Since AttnTemp~\cite{zou2024attention} noticed the attention collapse problem in directly transferring attention maps to target domains, we find that during the few-shot fine-tuning, MaPLe struggles to  rectify the collapse problem compared with LoRA-CLIP,  where the  visual  [CLS] token predominantly focuses on itself, 
causing the  observed  modality   misalignment  and  insufficient  class  separation  phenomenon  in Fig.\ref{fig:motivation}b. 
In other words, 
the model can hardly be fine-tuned to focus on the correct visual regions corresponding to the given class-related text, leading to 
the
modality  misalignment and bad class separation of visual samples, which finally causes the performance gap.  


To rectify attention collapse, we move beyond the structural parameter modifications of LoRA and seek to harness the efficacy of in-domain fine-tuning methods  
for  
CDFSL scenarios. Our observations reveal that the textual [EOS] token exhibits superior cross-attention toward class-relevant visual regions, acting as a \textit{semantic oracle} to steer the visual [CLS] token’s focus without requiring extensive parameter updates.
Moreover, we also find that CLIP’s standard InfoNCE loss
primarily enforces class separation yet provides weak constraints for modality alignment, consistent with 
the observations in 
\textcite{liang2022mind} and \textcite{tyshchuk2023isotropy}.
 Therefore, we propose a novel Semantic Probe framework, which comprises  two 
synergistic 
components: (1) an EOS-guided Attention Rectification (EAR) module to correct attention maps for better capturing the visual patterns corresponding to class-related texts, and (2) a 
dynamic  
Balanced Alignment and Separation (BAS) loss to 
enhance the learning of 
modality alignment and class separation. 

In summary, our contribution can be listed as:  
\begin{itemize} 
 \item{By establishing multiple fine-tuning baselines of CLIP on the CDFSL task, we are the first to find the performance gap between adapter-based 
 and prompt-based 
 methods, 
 opposite to the in-domain 
 scenarios. }
 \item{We delve into this phenomenon for  
 interpretation and find that the reason lies in the difference in the fine-tuning of attention networks, which causes the modality misalignment and  poor class separation problem, and finally leads to the performance gap.} 
  \item{Based on the interpretation, we further propose a Semantic Probe framework, which includes an EAR module to improve the fine-tuning of attention networks, and a BAS module to enhance 
 the learning of 
 both    
 modality alignment and class separation.} 
 \item{Extensive experiments on four widely used CDFSL datasets show that Semantic Probe consistently improves the performance of representative fine-tuning methods.}
 \end{itemize} 

\section{Why Does LoRA Show Better Fine-tuning Results in the CDFSL Task?}
Due to the 
 strong 
 performance shown in Fig.~\ref{fig:motivation}a,  we  select  LoRA-CLIP  \cite{hu2022lora, zanella2024low}  and MaPLe \cite{khattak2023maple} as the representative adapter-based and prompt-based methods,  respectively. 

\subsection{Preliminaries}
Cross-Domain Few-Shot Learning (CDFSL)
involves
adapting a model  to a target dataset $D_T = \{(x_i^T, y_i^T)\}_{i=1}^{N_T}$, where only scarce data is available, after pretraining  it on a source dataset $D_S = \{(x_i^S, y_i^S)\}_{i=1}^{N_S}$, which contains abundant labeled samples.
Typically, target datasets 
have large distribution  differences  from 
general domains, such as medical imaging or satellite imagery. CDFSL approaches commonly follow the \textit{N-way K-shot} paradigm to construct episodes (i.e., small datasets) from $D_T$. Each episode consists of a support set, denoted as $S = \{(x_{ij}, y_i)\}_{i=1,j=1}^{N,K}$, with $N$ 
classes and $K$ samples per class for training, and a  disjoint  query set, denoted as $Q = \{(x_{iq})\}_{i=1,q=1}^{N,M}$, $S \cap Q = \varnothing$, for evaluating performance on these $N$ classes.

CLIP 
introduces a contrastive learning framework
to align visual and textual modalities. For each 
image-text pair $(x_i^v, x_i^t)$, CLIP employs two independent encoders to extract 
features, $v_i = f_v(x_i^v)$ and $t_i = f_t(x_i^t)$, trained using a symmetric 
contrastive loss over a batch of $B$ positive pairs: 
\begin{equation}
L_{\text{clip}} = \frac{L_{i \to t} + L_{t \to i}}{2}
\label{eq:clip_full}
\end{equation}

MaPLe 
is the first multi-modal 
 prompt-tuning
method designed for CLIP.  
It jointly updates textual and visual prompts 
across layers
via a vision–language coupling function. 
At each transformer layer~$j$~($0 \le j < L$), the updates are defined as:
\begin{equation}
\begin{aligned}
\relax [P_j, W_j] &= T_j(P_{j-1}, W_{j-1}) \\
[c_j, E_j, \tilde{P}_j] &= V_j([c_{j-1}, E_{j-1}, F_{j-1}(P_{j-1})])
\end{aligned}
\end{equation}
where 
$P$ represents the textual prompt tokens, $W$ and $E$ denote the textual and visual input tokens, respectively, $T$ and $V$ are the text and image encoders, $F$ is the vision-language coupling function, 
and $c$ is the visual class token.

LoRA 
is  an adapter-based reparameterization  tuning approach \cite{DBLP:conf/emnlp/HuWLXLB0PL23} that enhances the efficiency of fine-tuning large-scale models by  injecting low-rank matrices into the attention layers of transformer architectures. Specifically, it modifies the weight matrices of the attention mechanism by adding low-rank updates (\(A_*, B_*\)), which significantly reduce the number of trainable parameters 
 while introducing no additional inference latency. 
The updates are applied as follows: 
\begin{equation}
    \begin{aligned}
        W_q^{'} &= W_q + \Delta W_q = W_q + A_qB_q\\
        W_k^{'} &= W_k + \Delta W_k = W_k + A_kB_k\\
        W_v^{'} &= W_v + \Delta W_v = W_v + A_vB_v\\
    \end{aligned}
\end{equation}

Additionally, to align with CLIP's pre-training paradigm and 
enrich textual modality information~\cite{schrodi2024two}, we 
 select 30 descriptions  per support set class as textual prompts during the fine-tuning phase of each episode.  The class-wise mean of text features is used as weights for the text classifier. 
Further details are provided in 
Section~\ref{prompt}.

\subsection{CLIP suffers from modality gap and insufficient class separation under CDFSL}
\begin{figure}
  \centering  
  \includegraphics[width=\linewidth]{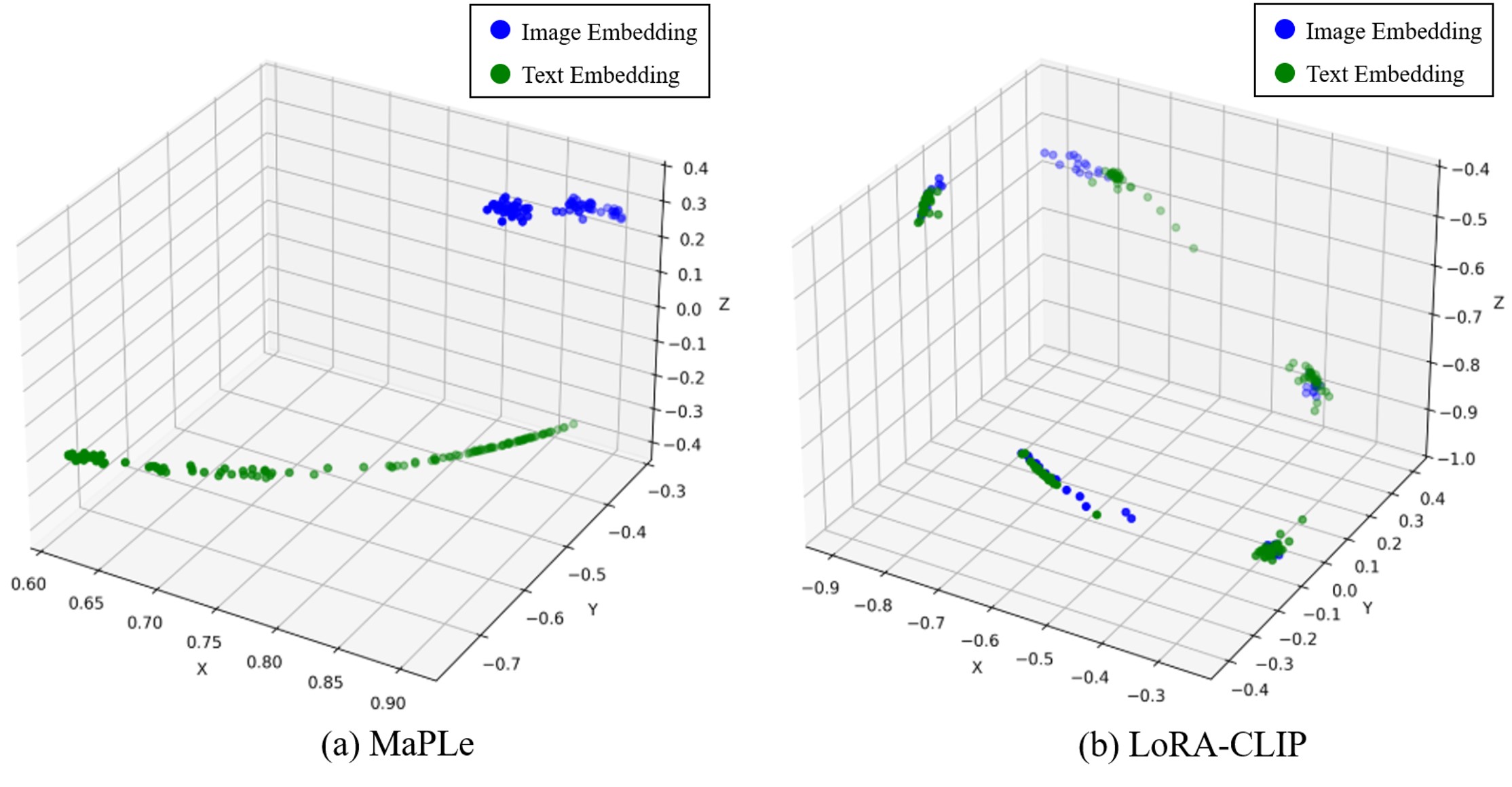}    \vspace{-0.45cm}  
  \caption{DOSNES visualization of embeddings of MaPLe and LoRA-CLIP. LoRA-CLIP (right) yields a smaller modality gap and tighter clusters.}  
  \label{fig:3d}  
\end{figure}

To investigate this performance gap, we first visualize the feature distributions of both image features (blue) and text features (green) extracted by LoRA-CLIP and MaPLe 
on the target domain, using DOSNES~\cite{lu2016doubly} projection, as shown in Fig.~\ref{fig:3d}. Our key observations reveal that: In cross-domain scenarios, (1) MaPLe exhibits 
pronounced 
modality gap between visual and textual features, with poor class separation; whereas (2) LoRA-CLIP projects both visual and textual features with less modality separation, while maintaining well-separated clusters. 

\begin{figure}
  \centering
  \includegraphics[width=\linewidth]{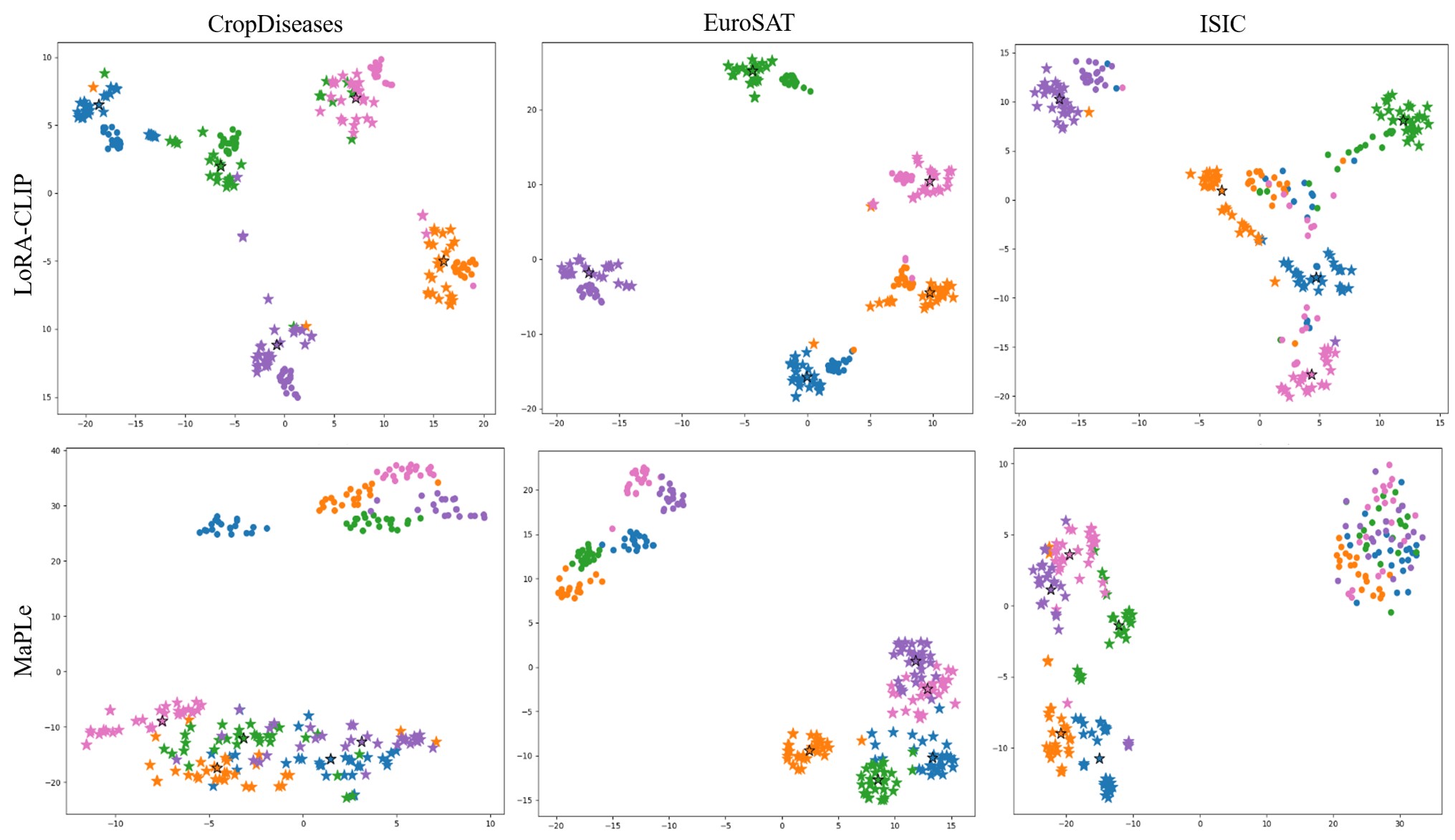}  \vspace{-0.3cm}
  \caption{T-SNE visualization of the embeddings of  LoRA-CLIP and MaPLe. The former (top  row) achieves clearer class boundaries for both visual  and textual features.} 
  \label{fig:2d}
\end{figure}

\begin{figure*}[t]
    \centering
    \includegraphics[width=\linewidth]{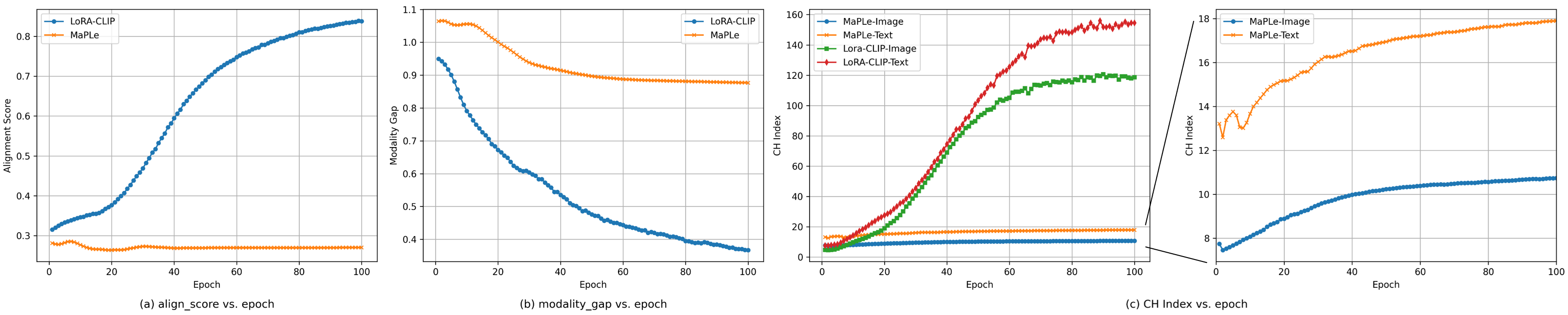}  \vspace{-0.45cm}
    \caption{
    The modality alignment and class separation metrics of the last-layer features evolve with training epochs for both LoRA-CLIP and MaPLe: 
    (a) align\_score (higher is better).  
(b) modality\_gap (lower is better).  
(c) Calinski-Harabasz Index (higher is better).  
    LoRA-CLIP is more effective than MaPLe at optimizing both modality alignment and class separation during fine-tuning for 
    CDFSL  
    tasks.
    }
    \label{fig:align+ch}
\end{figure*}
\begin{figure*}[thbp]
    \centering
    \includegraphics[width=\linewidth]{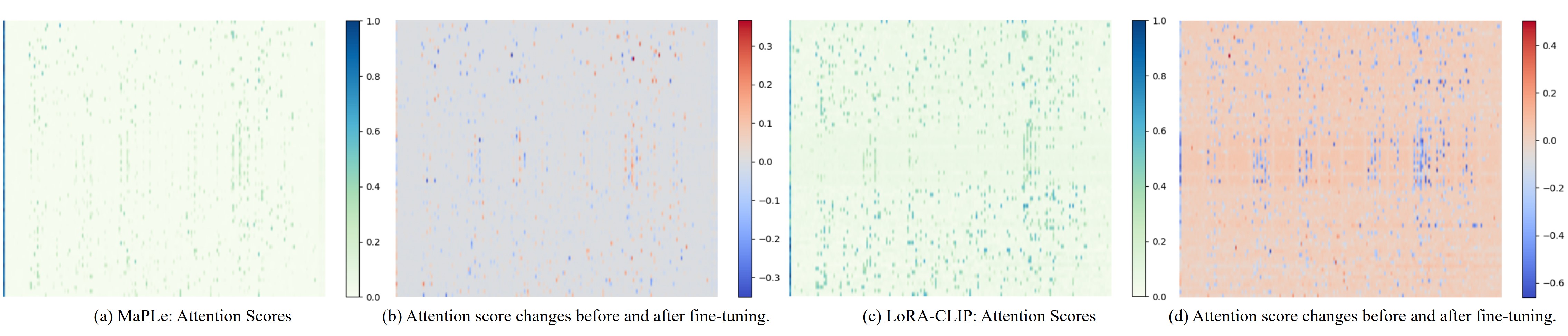}  \vspace{-0.45cm}
    \caption{
	Attention scores of the CLS token on image tokens and the changes after fine-tuning. 
    Before fine-tuning, the CLS token exhibits severe self-attention and neglects informative regions.  
    LoRA markedly shifts attention toward class-relevant patches after fine-tuning, while MaPLe yields almost no change.
    } %
    \label{fig:attention}
\end{figure*}

Fig.~\ref{fig:2d}  illustrates the distribution of class features by the T-SNE 
visualization. The text modality includes  30 text features 
and one class mean feature
 (marked with \textcolor{black}{\textbf{$\bigstar$}})
for each class, 
while image features are represented by \scalebox{0.8}{\(\bigcirc\)},   with different classes distinguished by different colors. In comparison to  MaPLe, LoRA-CLIP demonstrates superior class separability, which is advantageous for classification tasks.

To further quantitatively validate our observations, we employ   the 
 \textit{$align\_score$} \cite{eslami2024mitigate} and \textit{$modality\_gap$} \cite{liang2022mind} to assess modality alignment, alongside the Calinski-Harabasz (CH) 
 Index to evaluate class separability. The \textit{$align\_score$} is defined as follows:
 \vspace{-0.1cm}
\begin{equation}
    align\_score = \frac{1}{U\times |S|}\sum_{i\in(0,N)}{e_{i}^{v}\cdot e_{i}^{t}}
\end{equation}
where $e^{v}$ represents  image features, $e^{t}$ represents the text features corresponding to the true class, $U$ (i.e., 30) denotes  the number of textual prompts per class,  \( |S| = N \times K \) is the size of the support set, and $N$ is the number of classes. 

The \textit{modality\_gap}  quantifies the Euclidean distance between the mean vectors 
of image and text features, defined as: 
\begin{equation}
    modality\_gap = \left\| \frac{1}{|S|} \sum_{i=1}^{|S|} e_i^v - \frac{1}{N_t} \sum_{i=1}^{N_t} e_i^t \right\|_2
\end{equation}
 where $N_t  = N \times U$ is the total number of text prompts  across all classes. 

The CH Index evaluates the ratio of inter-class compactness to intra-class separation, calculated as:
{
\begin{equation}
    CH(X, L) = \frac{B(X,L)\cdot (n_{samples}-n_{labels})}{W(X,L)\cdot (n_{labels}-1)}
\end{equation}
}
where  $X$ represents  the   feature matrix, containing $n_{samples}$ data points, $L$ represents the cluster labels, containing $n_{labels}$ categories, 
$B(X,L)$ represents the between-cluster dispersion, and $W(X,L)$ denotes the within-cluster dispersion. 
A higher CH index value indicates greater class separation. 

Fig.~\ref{fig:align+ch}(a,b) 
shows 
the modality alignment metrics from the final layer outputs of LoRA-CLIP and MaPLe across training epochs. 
 In Fig.~\ref{fig:align+ch}a, the \textit{align\_score} 
 (higher values indicate better alignment) 
 demonstrates that LoRA-CLIP consistently outperforms MaPLe. Similarly, Fig.~\ref{fig:align+ch}b illustrates the \textit{modality\_gap} 
 (lower values are preferable), 
 where LoRA-CLIP achieves a substantially smaller gap compared to MaPLe. Fig.~\ref{fig:align+ch}c depicts the CH Index for both visual and textual features over training epochs, revealing an increasing trend for both models, with LoRA-CLIP converging to a significantly higher value than MaPLe, indicating superior class separability.

In summary, the above results demonstrate that \textbf{LoRA-CLIP consistently
outperforms MaPLe in both modality alignment and class separation for CDFSL tasks}, 
which are closely related to their performance disparities.

\subsection{Collapsed attention under CDFSL leads to modality gap and insufficient class separation}                   
\label{sec2.3}

Prior work~\cite{zou2024attention} has shown that visual attention transferred to cross-domain tasks, particularly distant-domain tasks, tends to collapse toward a small number of tokens. Following its L1-based evaluation of attention density, we adopt the same metric and adapt it to the CLIP fine-tuning methods considered in this study. Specifically, we compute the mean L1 norm of the final-layer visual self-attention after row-wise L2 normalization. The calculation includes image-token queries and patch-token keys while excluding learned prompt tokens and other auxiliary representation tokens. A lower L1 density indicates a sparser and more concentrated attention distribution and, therefore, more severe attention collapse.

\begin{table}
\centering
\caption{Final-layer visual-attention L1 density under representative fine-tuning methods. 
}
\label{tab:l1_density}
\setlength{\tabcolsep}{7pt}
\resizebox{0.48\textwidth}{!}{
\begin{tabular}{llccccc}
\toprule
Method & Type & ChestX & ISIC & EuroSAT & CropDiseases & Average \\
\midrule
MaPLe
& Prompt 
& 11.3107
& 10.8601
& 10.9581
& 11.1397
& 11.0672 \\
LoRA-CLIP
& Adapter 
& 11.1719
& 10.9704
& 11.2597
& 11.2153
& 11.1543 \\
MMRL
& Prompt 
& 11.2510
& 11.1771
& 11.1934
& 11.1684
& 11.1975 \\
MMA
& Adapter 
& 11.6344
& 11.2359
& 11.8918
& 11.5376
& 11.5749 \\
\bottomrule
\end{tabular}
}
\vspace{-0.3cm} 
\end{table}
As shown in Table~\ref{tab:l1_density}, LoRA-CLIP achieves a higher average L1 density than MaPLe, while MMA~\cite{Yang_2024MMA_CVPR} similarly achieves a higher average L1 density than MMRL~\cite{DBLP:conf/cvpr/GuoG25}. All methods are evaluated using identical class-name textual information and the same 5-way 5-shot protocol. These matched comparisons indicate that the adapter-based methods exhibit less severe attention collapse than their prompt-based counterparts, consistent with the performance advantages of LoRA-CLIP over MaPLe and MMA over MMRL.

Since 
\textcite{zou2024a}
 pointed out that the attention \textit{directly transferred} to cross domains (especially distant domains) tends to collapse (focusing solely on the CLS token),
we are inspired to analyze the attention distribution of the visual CLS token under CLIP to understand modality gap and class separation issues.
As shown in Fig.~\ref{fig:attention}, 
the Y-axis corresponds to the batch dimension, and the X-axis represents the token positions within each sample.
Fig.~\ref{fig:attention}a shows the heatmap of the CLS token’s attention distribution over  image tokens, with Fig.~\ref{fig:attention}b  depicting changes in CLS 
token’s 
attention scores before and after fine-tuning.  Fig.~\ref{fig:attention}c and Fig.~\ref{fig:attention}d present the corresponding attention heatmaps for LoRA-CLIP arranged in the same order.  

Fig.\ref{fig:attention} highlights two key findings in the visual features:

$\bullet$ 
\textbf{Attention Collapse:} 
An attention collapse problem exists widely, 
consistent with  the findings of 
\textcite{zou2024attention},   
where the CLS token predominantly focuses on itself across samples, as seen in the high attention scores in the leftmost columns of 
Fig.~\ref{fig:attention}a and Fig.~\ref{fig:attention}c.

\begin{figure*}
\centering  
\subfloat[Heatmaps of CLS and EOS attention to local image tokens] 
{
\includegraphics[width=0.53\linewidth]{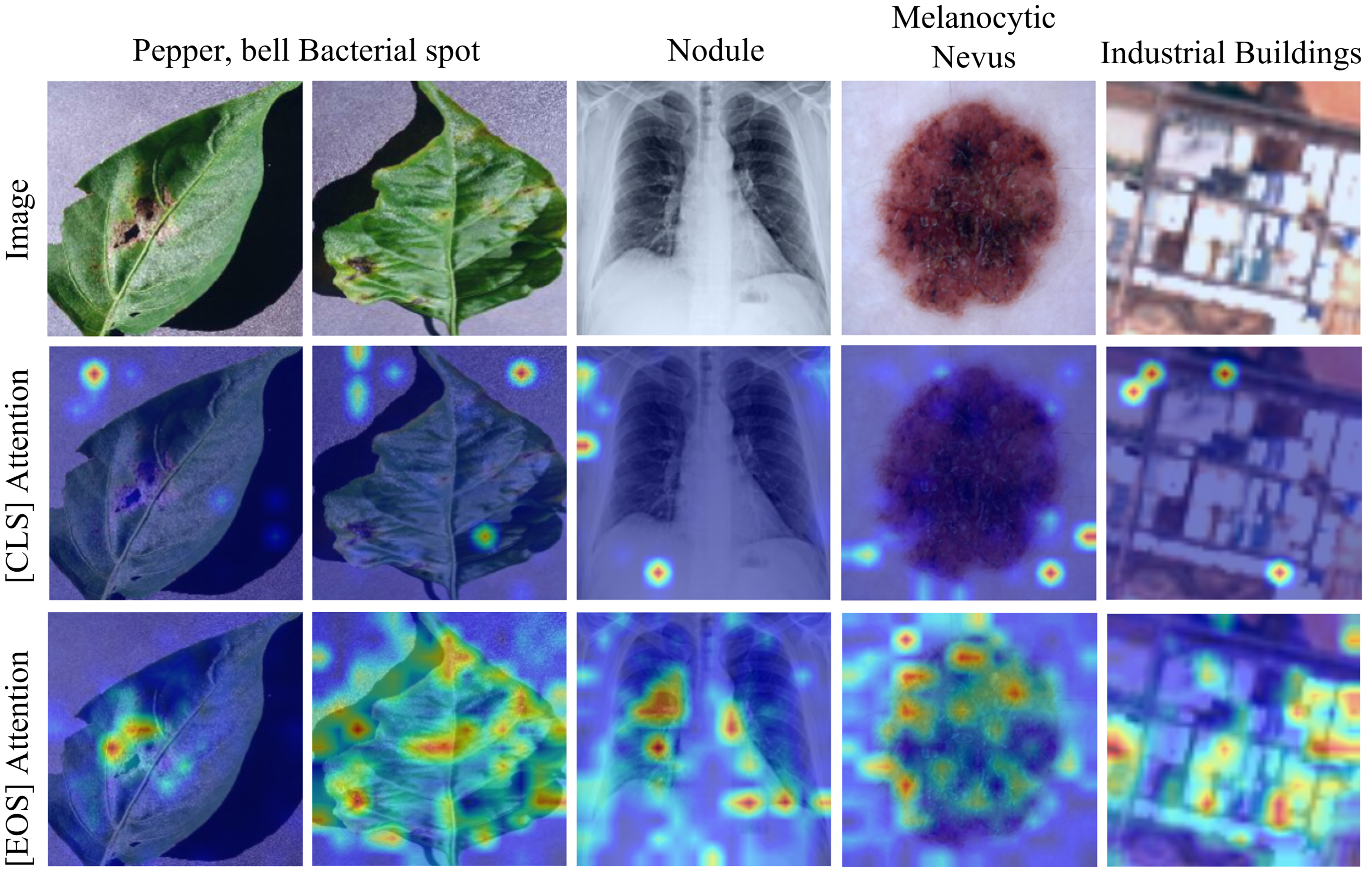}%
\label{fig:eoscls111}
}
\hfil
\subfloat[ EOS-guided CLS Attention Rectification]
{
\includegraphics[width=0.44\linewidth]{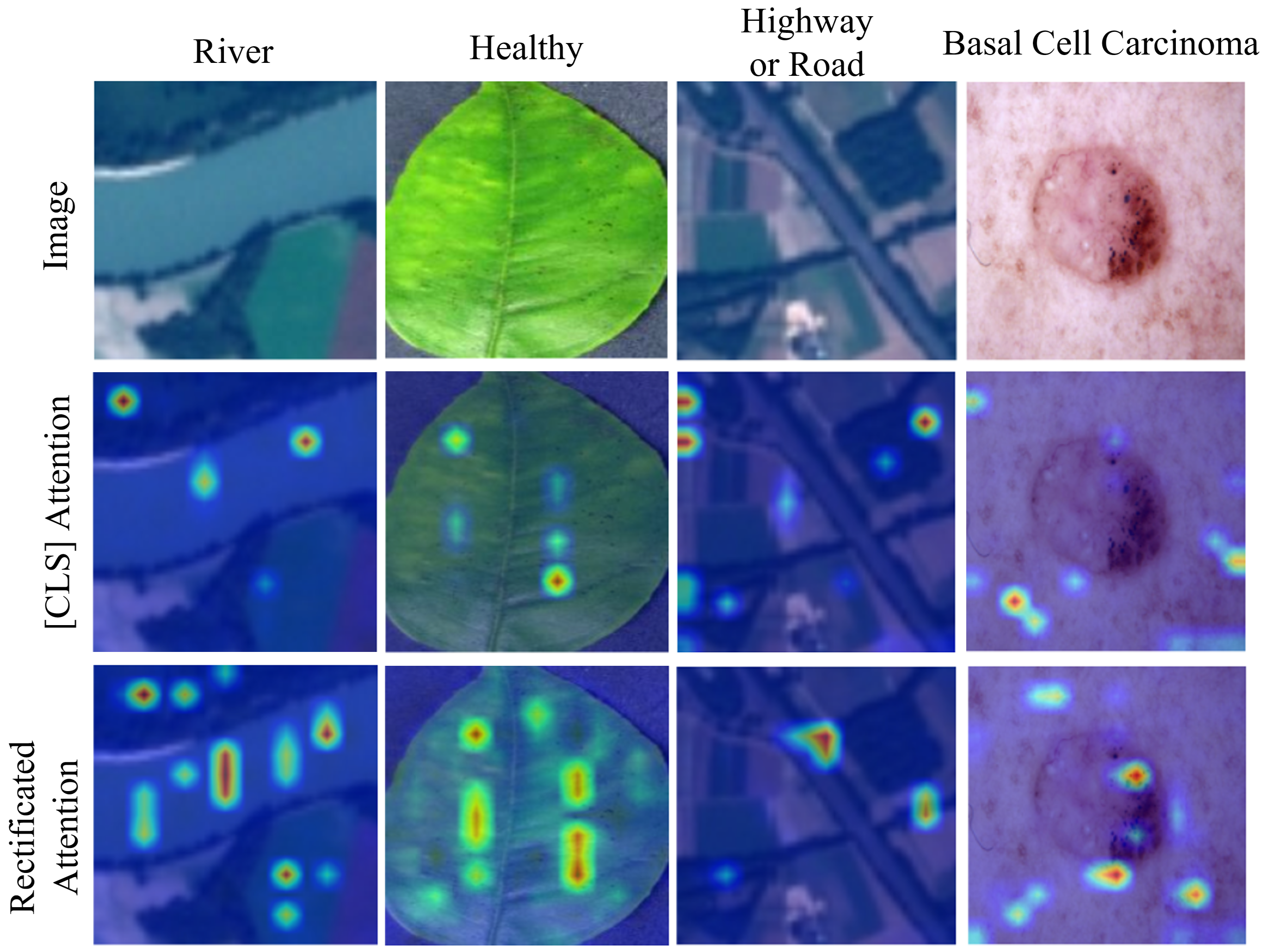}%
   \label{fig:eoscls222}
}       
\caption
{
  (a) Heatmaps of the visual CLS and textual EOS token's attention to image. 
  The EOS token demonstrates significantly stronger capability in capturing visual semantic information compared to the CLS token. (b) EOS-guided CLS Attention Rectification. 
}
\label{fig:eoscls} 
\end{figure*} 

$\bullet$ 
\textbf{LoRA-CLIP vs. MaPLe:}  
MaPLe struggles to address  the attention collapse issue, as shown in Fig.~\ref{fig:attention}b, where most attention scores remain unchanged  
(gray means zero change) after the  
cross-domain 
fine-tuning, and the major attention is still on the CLS token in Fig.~\ref{fig:attention}a. Conversely, Fig.~\ref{fig:attention}d demonstrates  that LoRA can effectively adjust  the attention scores during  fine-tuning, resulting in diverse focus across various local patch tokens in Fig.~\ref{fig:attention}c.

Therefore, we hold that this inconsistency in attention distribution 
leads to 
both the modality gap and the poor class separation problem: 

$\bullet$ 
\textbf{Modality Misalignment:}  
Given the text that describes the visual clues in the image, if the visual CLS token  fails to focus on the corresponding regions of the image, the visual feature can hardly match the text feature, leading to modality misalignment and gap.

$\bullet$ 
\textbf{Poor Class Separation:} 
If the CLS token can't focus on discriminative patches in the image, the discriminability of the visual feature will also be harmed, 
leading to poor
class separation.

To directly examine whether attention collapse affects modality alignment and class separation, we conduct a controlled noise intervention on the original LoRA-CLIP and MaPLe models without Semantic Probe. All training and evaluation settings are kept fixed within each method, and only the noise strength applied to the last two visual-attention layers is varied on miniImageNet. As shown in Table~\ref{tab:attention_collapse_intervention}, increasing the noise strength consistently decreases the attention entropy and increases both the maximum attention mass and inter-sample attention similarity, indicating progressively stronger and more homogeneous attention collapse. Meanwhile, both the alignment score and the Image CH index decrease monotonically, demonstrating that stronger attention collapse impairs image-text alignment and visual class aggregation. These controlled results provide direct evidence that attention collapse  
leads to 
 the modality gap and insufficient class separation in cross-domain few-shot learning.

\begin{table}
\centering
\caption{Controlled attention-collapse intervention on miniImageNet. 
}
\label{tab:attention_collapse_intervention}
\setlength{\tabcolsep}{6pt}
\resizebox{0.48\textwidth}{!}{
\begin{tabular}{lcccccc}
\toprule
Method
& Noise
& Entropy
& Maximum mass
& Attention similarity
& Align score
& Image CH \\
\midrule
\multirow{5}{*}{LoRA-CLIP}
& 0 & 0.7415 & 0.1321 & 0.1343 & 0.2837 & 6.27 \\
& 1 & 0.6670 & 0.2215 & 0.1667 & 0.2806 & 5.89 \\
& 3 & 0.4501 & 0.3578 & 0.3600 & 0.2729 & 4.51 \\
& 5 & 0.3452 & 0.4321 & 0.4894 & 0.2670 & 3.85 \\
& 7 & 0.2883 & 0.4840 & 0.5588 & 0.2619 & 3.51 \\
\midrule
\multirow{5}{*}{MaPLe}
& 0 & 0.7374 & 0.1341 & 0.1331 & 0.2756 & 4.90 \\
& 1 & 0.6618 & 0.2261 & 0.1660 & 0.2740 & 4.67 \\
& 3 & 0.4450 & 0.3664 & 0.3562 & 0.2662 & 3.63 \\
& 5 & 0.3427 & 0.4377 & 0.4839 & 0.2615 & 3.09 \\
& 7 & 0.2856 & 0.4920 & 0.5508 & 0.2544 & 2.76 \\
\bottomrule
\end{tabular}
}
\vspace{-0.35cm}
\end{table}

Based on the above observations, we can explain the superiority of  LoRA over other fine-tuning methods in cross-domain scenarios  as follows: LoRA directly modifies the parameter matrices (i.e., the $Q/K/V$ parameters) involved in the 
attention mechanism, enabling a more effective correction of the collapsed attention.
This, in turn, helps mitigate the 
resulting issues of modality misalignment and poor class separation, thereby 
significantly improving CLIP's performance in CDFSL tasks.

\subsection{Inspiration: addressing attention collapse by introducing text information}
Since we verified that the rectification of collapsed attention is the core of effective few-shot fine-tuning in the CDFSL problem, a question naturally arises:  \textbf{\textit{Can we use another way (other than LoRA) to help the attention rectification to better utilize those advantageous in-domain fine-tuning methods in CDFSL?}} Given the text modality, we ask: can we take the text information to improve the rectification? 
While current work~\cite{zou2024attention} merely focuses on the CLS token's  attention in solely the visual branch,  the text modality actually provides prior knowledge about what the vision network should focus on, 
potentially guiding the calibration of the CLS token's attention.

For example, when reading an unfamiliar article, 
we may easily 
overlook
important information in this article, just like how the visual CLS token misses critical regions 
of  
 the image in 
cross-domain scenarios. However, if  an oracle provides us with the information about what to focus on (such as finding the birthday of someone), it will be easier for us to capture 
the relevant details.
 In this context, the text modality essentially serves as the oracle,
directing the vision network to relevant regions and aiding attention rectification. 

To validate this intuition, 
we further visualize  the heatmaps of the CLS token and the textual EOS token's  attention to 
local patch tokens in Fig.\ref{fig:eoscls}a. We find that the EOS token consistently focuses on 
more accurate and comprehensive discriminative regions compared with the visual CLS token, which 
verifies our intuition and motivates us to 
 leverage the EOS token 
 to enhance attention rectification in CLIP for CDFSL tasks. 
Based on this intuition, we can further enhance the attention rectification beyond  LoRA, reviving those  
in-domain fine-tuning methods 
for CDFSL. 

\begin{figure*}[t]
\centering
\subfloat[]
{
\includegraphics[width=0.49\linewidth]{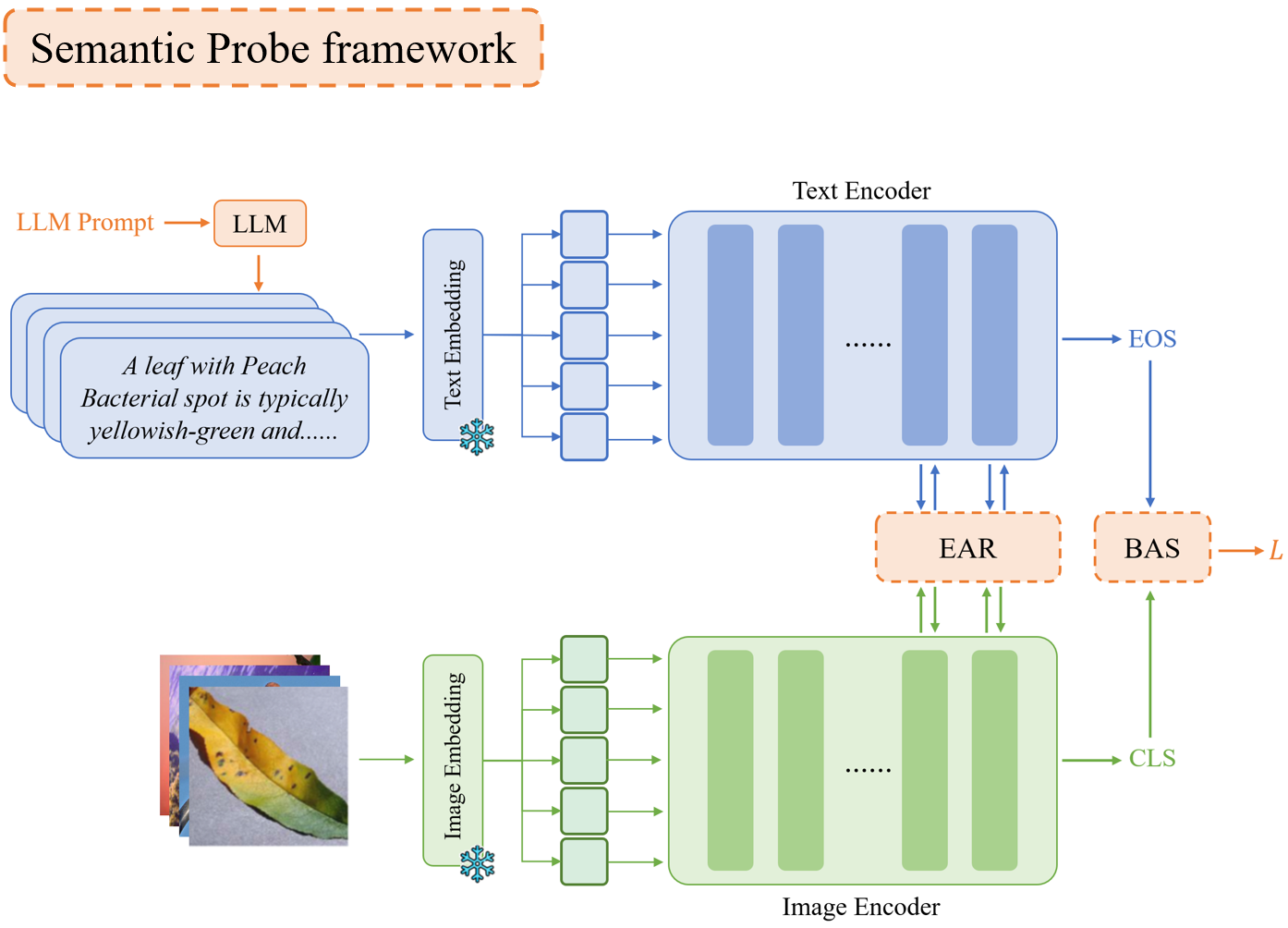}%
\label{fig:framework}
}
\hfil
\subfloat[]
{
\includegraphics[width=0.48\linewidth]{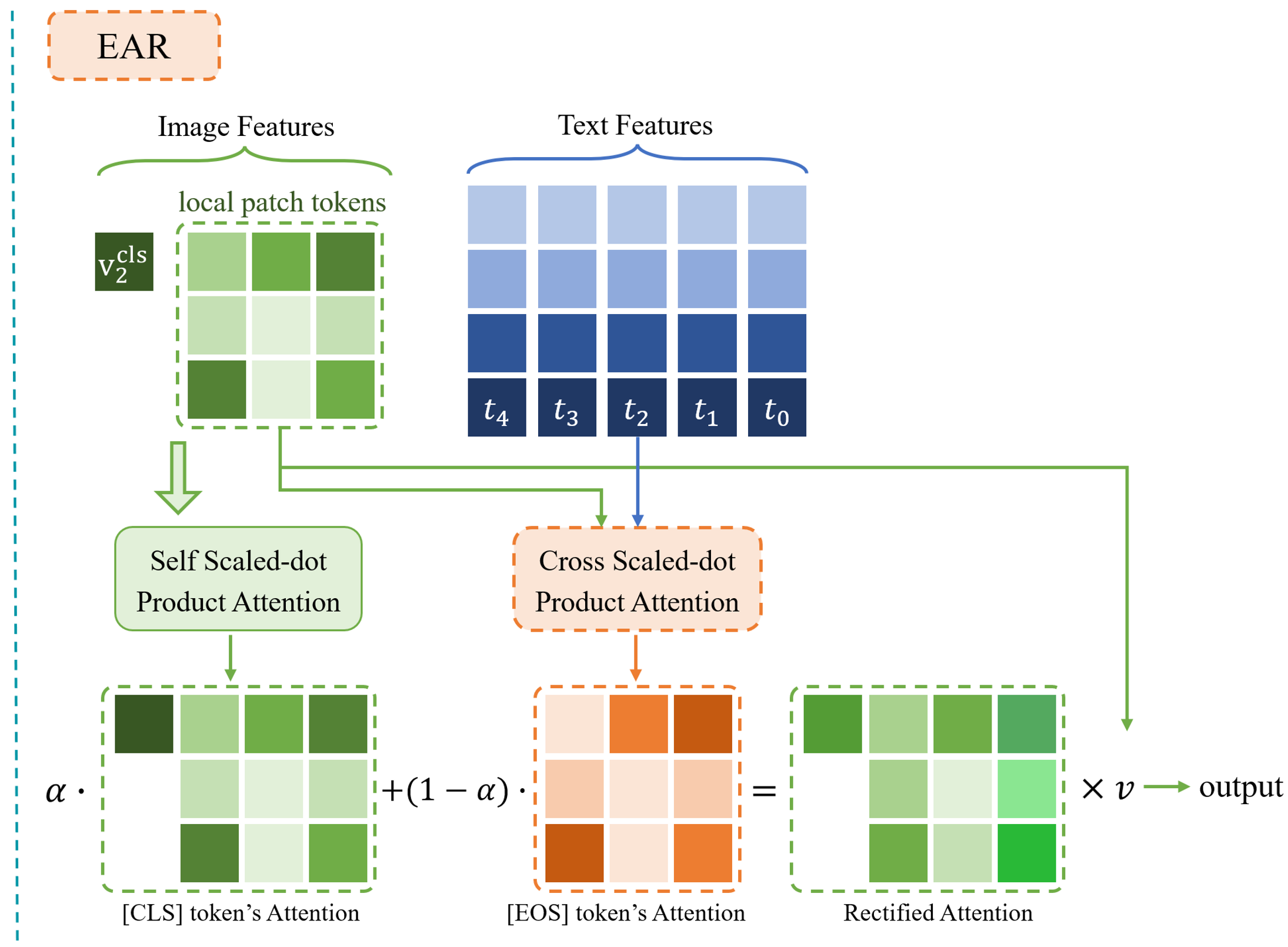}%
   \label{fig:ear}
}    
\caption
{
(a) Overview of our 
 Semantic Probe framework, which  revives in-domain fine-tuning methods for
 CDFSL. 
 The EOS-guided Attention Rectification 
 module is 
  plugged 
 to the final layers of CLIP to rectify attention, while the Balanced Alignment and Separation 
 loss replaces the original contrastive loss to dynamically guide the model’s focus between modality alignment and class separation. (b) EAR module. The attention matrix from the EOS token to visual features is used to rectify the attention of the CLS token. Through weighted summation, EAR reduces the CLS token’s excessive self-focus and redirects its attention toward discriminative visual tokens.
}  
\end{figure*}

\section{Methods}
Based on the above  
interpretation and 
analysis, we aim to address the attention collapse problem and improve both the modality alignment and class separation.
To address these issues, we propose the Semantic Probe framework, which consists of two synergistic components: (1) an EOS-guided Attention Rectification (EAR) module to correct attention maps;  and (2) a Balanced Alignment and Separation (BAS) loss to dynamically enhance the learning of both the modality alignment and class separation, as illustrated in Fig.\ref{fig:framework}.

\subsection{EOS-guided Attention Rectification}
To correct the attention bias of the visual  CLS token in CLIP, we aim to enhance its focus on semantically informative image tokens. These tokens can be identified using the textual EOS token’s attention, which inherently captures richer class-specific information. As shown in Fig.~\ref{fig:eoscls}b, when the EOS attention matrix is used to refine the CLS attention, the CLS token exhibits improved focus on relevant image regions and becomes more aligned with texts.

Building upon this, we propose an EOS-guided Attention Rectification module (Fig.~\ref{fig:ear}). 
For the i-th layer in CLIP,  the attention weights of the CLS token with respect to all visual 
tokens  are defined as: 

\vspace{-0.45cm}
{
\small 
\begin{equation}
    Attention_{i}^{cls} = softmax(\frac{v^{cls}\cdot W_{q}(v\cdot W_k)^T}{\sqrt{d_v}}) \in \mathbb{R}^{1\times(1+n)} 
\label{eq:cls}
\end{equation}
}
Here, $ v^{cls} = v[0] \in \mathbb{R}^{d_v}$ represents the global CLS token, $v\in \mathbb{R}^{(1+n) \times d_v}$ denotes the sequence of visual tokens input to this layer (including both the CLS token and image patch tokens), and $ W_{q} $ , $ W_{k} $ $\in\mathbb{R}^{d_v\times d_v}$ are the 
learnable 
Query and Key transformation matrices, respectively.

  Concurrently, we compute the attention matrix of the textual EOS token  over the $n$ image patch tokens:   
  {
\small
\begin{equation}
    Attention_{i}^{eos} = softmax( \frac{ (t^{eos} \cdot E) \cdot W_{q}^{'}(v[1:]\cdot W_k^{'})^T}{\sqrt{d_v}})  
    \label{eq:eos}
\end{equation}
}
In this equation, 
$t^{eos} \in \mathbb{R}^{d_t}$  represents the  
textual 
EOS token  obtained from the i-th layer of CLIP’s text encoder, $v[1:] \in \mathbb{R}^{n \times d_v}$ specifically refers to the sequence of image patch tokens, excluding the visual CLS token, as our focus is on the EOS token's attention to the content regions of the image. 
Due to the inherent dimensional discrepancy between visual and textual 
embeddings, 
 the  
 $ E $ 
 matrix is designed with dimensions $\mathbb{R}^{d_t\times d_v}$ 
  to project the textual EOS token into the visual feature space, 
 enabling effective cross-modal interaction.
Note that the resulting EOS-to-image guidance is an explicitly constructed cross-modal attention module rather than an attention mechanism native to CLIP.  

Subsequently, we  calculate the rectified attention distribution matrix  
by performing a weighted summation of the above two attention matrices:
{ 
\small 
\begin{equation} 
    Attention_{i}^{cls^{'}} = \alpha \cdot Attention_{i}^{cls} + (1-\alpha)\cdot [0;Attention_{i}^{eos}] 
\label{eqq:alpha} 
\end{equation} 
} 
where $\alpha$ is a coefficient with a value less than 1. After multiplying $Attention_{i}^{cls}$ by $\alpha$, we  effectively reduce  the CLS token's self-attention score, thereby suppressing its tendency for excessive self-focus. 
As demonstrated in Section~\ref{sec2.3}, the attention scores for image tokens within  $Attention_{i}^{eos}\in \mathbb{R}^{1 \times n} $
  are typically higher than their corresponding values in $Attention_{i}^{cls} $.
 Consequently, this weighted summation increases the attention scores directed towards the image tokens, guiding the global CLS token to concentrate more on discriminative, category-relevant visual information. It is important to note that since both attention distributions are normalized via the softmax function (i.e., 
$\sum Attention_{i}^{eos} = \sum Attention_{i}^{cls} = 1$),
their weighted sum also remains normalized, ensuring the stability of the feature distribution.
In practice, to achieve dimension alignment, we pad a zero before the elements of   
$Attention_{i}^{eos}$, making its dimension consistent with $Attention_{i}^{cls}$. 


\newcommand{\accCI}[2]{\ensuremath{#1\,{\scriptstyle\pm}\,#2}} 

\begin{table*}[th]
\centering
\setlength{\abovecaptionskip}{2pt}
\caption{Accuracies (\%) of target domain datasets of 5-way 1-shot and 5-shot tasks. 
 $^*$ denotes that data augmentation was utilized in our implementation.  
 Refer to the Appendix for the extended table. }
\label{tab:main_results}
\begin{adjustbox}{max width=0.95\textwidth}
\begin{threeparttable}
\arrayrulecolor{black}
\arrayrulewidth=0.6pt
\begin{tabular}{l|l|c|c|cccc|c|}
\hline
\textbf{Task}
& \textbf{Method}
& \textbf{Mark}
& \textbf{backbone}
& \textbf{ISIC}
& \textbf{ChestX}
& \textbf{EuroSAT}
& \textbf{CropDisease}
& \textbf{Avg} \\
\hline

\multirow{16}{*}{\rotatebox{90}{\textbf{5-way 1-shot}}}
& StepSPT~\cite{DBLP:journals/corr/abs-2411-10070}
& TPAMI-25 & ViT/CLIP
& \accCI{32.97}{0.27}
& \accCI{22.84}{0.95}
& \accCI{70.01}{0.21}
& \accCI{84.84}{0.72}
& 52.66 \\

& Tip-Adapter$^*$~\cite{Tip-Adapter}
& ECCV-22 & ViT/CLIP
& \accCI{32.68}{0.37}
& \accCI{22.24}{0.26}
& \accCI{75.44}{0.51}
& \accCI{77.15}{0.66}
& 51.87 \\

& AMU-Tuning$^*$~\cite{DBLP:conf/cvpr/TangLW0H24}
& CVPR-24 & ViT/CLIP
& \accCI{32.29}{0.67}
& \accCI{21.56}{0.36}
& \accCI{72.24}{0.71}
& \accCI{80.20}{0.86}
& 51.57 \\

& LP++$^*$~\cite{lp2024}
& CVPR-24 & ViT/CLIP
& \accCI{33.63}{0.41}
& \accCI{21.72}{0.42}
& \accCI{73.05}{0.55}
& \accCI{81.84}{0.66}
& 52.56 \\

& LDC$^*$~\cite{Li_2025_CVPR}
& CVPR-25 & ViT/CLIP
& \accCI{33.72}{0.46}
& \accCI{22.32}{0.36}
& \accCI{74.39}{0.52}
& \accCI{84.07}{0.61}
& 53.62 \\

& LoRA-CLIP~\cite{zanella2024low}
& CVPR-24 & RN50/CLIP
& \accCI{32.01}{0.62}
& \accCI{21.76}{0.42}
& \accCI{57.79}{1.08}
& \accCI{65.24}{1.01}
& 44.20 \\

& CoOp~\cite{zhou2022learning}
& IJCV-22 & ViT/CLIP
& \accCI{29.47}{0.67}
& \accCI{20.95}{0.44}
& \accCI{68.16}{1.00}
& \accCI{79.27}{1.06}
& 49.46 \\

& \cellcolor{cyan!10}\textbf{CoOp + OURS}
& \cellcolor{cyan!10}-
& \cellcolor{cyan!10}ViT/CLIP
& \cellcolor{cyan!10}\accCI{29.64}{0.64}
& \cellcolor{cyan!10}\accCI{21.33}{0.43}
& \cellcolor{cyan!10}\accCI{69.77}{0.93}
& \cellcolor{cyan!10}\accCI{80.93}{1.03}
& \cellcolor{cyan!10}50.42 \\

& CLIP-Adapter~\cite{gao2024clip}
& IJCV-24 & ViT/CLIP
& \accCI{31.31}{0.71}
& \accCI{21.51}{0.47}
& \accCI{67.83}{0.97}
& \accCI{81.14}{1.06}
& 50.44 \\

& \cellcolor{cyan!10}\textbf{CLIP-Adapter + OURS}
& \cellcolor{cyan!10}-
& \cellcolor{cyan!10}ViT/CLIP
& \cellcolor{cyan!10}\accCI{32.83}{0.65}
& \cellcolor{cyan!10}\accCI{22.73}{0.36}
& \cellcolor{cyan!10}\accCI{70.19}{0.84}
& \cellcolor{cyan!10}\accCI{82.03}{1.01}
& \cellcolor{cyan!10}51.94 \\

& Tip-Adapter~\cite{Tip-Adapter}
& ECCV-22 & ViT/CLIP
& \accCI{30.36}{0.58}
& \accCI{21.96}{0.34}
& \accCI{72.05}{0.92}
& \accCI{76.55}{1.21}
& 50.23 \\

& \cellcolor{cyan!10}\textbf{Tip-Adapter + OURS}
& \cellcolor{cyan!10}-
& \cellcolor{cyan!10}ViT/CLIP
& \cellcolor{cyan!10}\accCI{31.88}{0.59}
& \cellcolor{cyan!10}\accCI{22.23}{0.40} 
& \cellcolor{cyan!10}\accCI{74.29}{0.96}
& \cellcolor{cyan!10}\accCI{78.19}{1.08}
& \cellcolor{cyan!10}51.65 \\

& MaPLe~\cite{khattak2023maple}
& CVPR-23 & ViT/CLIP
& \accCI{31.10}{0.70}
& \accCI{20.89}{0.44}
& \accCI{75.66}{1.03}
& \accCI{81.87}{1.00}
& 53.04 \\

& \cellcolor{cyan!10}\textbf{MaPLe + OURS} 
& \cellcolor{cyan!10}- 
& \cellcolor{cyan!10}ViT/CLIP 
& \cellcolor{cyan!10}\accCI{34.27}{0.74} 
& \cellcolor{cyan!10}\accCI{21.41}{0.40} 
& \cellcolor{cyan!10}\accCI{82.94}{0.87} 
& \cellcolor{cyan!10}\accCI{82.83}{1.01} 
& \cellcolor{cyan!10}55.36 \\ 

& LoRA-CLIP~\cite{zanella2024low}
& CVPR-24 & ViT/CLIP
& \accCI{34.45}{0.74}
& \accCI{21.73}{0.43}
& \accCI{80.34}{0.92}
& \accCI{84.95}{0.89}
& 55.37 \\

& \cellcolor{cyan!10}\textbf{LoRA-CLIP + OURS}
& \cellcolor{cyan!10}-
& \cellcolor{cyan!10}ViT/CLIP
& \cellcolor{cyan!10}\accCI{\textbf{38.77}}{0.78}
& \cellcolor{cyan!10}\accCI{\textbf{23.65}}{0.40}
& \cellcolor{cyan!10}\accCI{\textbf{82.94}}{0.77}
& \cellcolor{cyan!10}\accCI{\textbf{85.11}}{0.96}
& \cellcolor{cyan!10}\textbf{57.62} \\

\hline
\hline

\multirow{16}{*}{\rotatebox{90}{\textbf{5-way 5-shot}}}
& StepSPT~\cite{DBLP:journals/corr/abs-2411-10070}
& TPAMI-25 & ViT/CLIP
& \accCI{52.12}{0.36}
& \accCI{\textbf{26.36}}{0.97}
& \accCI{89.40}{1.05}
& \accCI{96.01}{0.88}
& 65.97 \\

& Tip-Adapter$^*$~\cite{Tip-Adapter}
& ECCV-22 & ViT/CLIP
& \accCI{46.96}{0.59}
& \accCI{24.07}{0.44}
& \accCI{87.24}{0.33}
& \accCI{94.19}{0.39}
& 63.12 \\

& AMU-Tuning$^*$~\cite{DBLP:conf/cvpr/TangLW0H24}
& CVPR-24 & ViT/CLIP
& \accCI{44.60}{0.62}
& \accCI{23.34}{0.41}
& \accCI{88.47}{0.39}
& \accCI{94.26}{0.52}
& 62.66 \\

& LP++$^*$~\cite{lp2024}
& CVPR-24 & ViT/CLIP
& \accCI{48.49}{0.44}
& \accCI{23.89}{0.29}
& \accCI{87.48}{0.42}
& \accCI{94.47}{0.38}
& 63.58 \\

& LDC$^*$~\cite{Li_2025_CVPR}
& CVPR-25 & ViT/CLIP
& \accCI{49.70}{0.33}
& \accCI{25.89}{0.21}
& \accCI{90.82}{0.22}
& \accCI{96.71}{0.34}
& 65.78 \\

& LoRA-CLIP~\cite{zanella2024low}\phantom{zhengzaipao}
& CVPR-24 & RN50/CLIP
& \accCI{46.53}{0.70}
& \accCI{23.00}{0.48}
& \accCI{77.26}{0.87}
& \accCI{87.42}{0.73}
& 58.55 \\

& CoOp~\cite{zhou2022learning}
& IJCV-22 & ViT/CLIP
& \accCI{42.56}{0.66}
& \accCI{22.22}{0.45}
& \accCI{84.62}{0.51}
& \accCI{92.81}{0.51}
& 60.55 \\

& \cellcolor{cyan!10}\textbf{CoOp + OURS}
& \cellcolor{cyan!10}-
& \cellcolor{cyan!10}ViT/CLIP 
& \cellcolor{cyan!10}\accCI{43.32}{0.70} 
& \cellcolor{cyan!10}\accCI{22.68}{0.46} 
& \cellcolor{cyan!10}\accCI{85.77}{0.52} 
& \cellcolor{cyan!10}\accCI{94.29}{0.57} 
& \cellcolor{cyan!10}61.52 \\

& CLIP-Adapter~\cite{gao2024clip}
& IJCV-24 & ViT/CLIP
& \accCI{44.09}{0.76}
& \accCI{23.56}{0.49}
& \accCI{84.48}{0.55}
& \accCI{93.34}{0.62}
& 61.36 \\

& \cellcolor{cyan!10}\textbf{CLIP-Adapter + OURS}
& \cellcolor{cyan!10}-
& \cellcolor{cyan!10}ViT/CLIP
& \cellcolor{cyan!10}\accCI{44.91}{0.73}
& \cellcolor{cyan!10}\accCI{23.73}{0.40}
& \cellcolor{cyan!10}\accCI{85.37}{0.52}
& \cellcolor{cyan!10}\accCI{93.00}{0.55}
& \cellcolor{cyan!10}61.75 \\

& Tip-Adapter~\cite{Tip-Adapter}
& ECCV-22 & ViT/CLIP
& \accCI{44.12}{0.67}
& \accCI{23.82}{0.47}
& \accCI{84.48}{0.54}
& \accCI{92.63}{0.54}
& 61.26 \\ 

& \cellcolor{cyan!10}\textbf{Tip-Adapter + OURS} 
& \cellcolor{cyan!10}- 
& \cellcolor{cyan!10}ViT/CLIP 
& \cellcolor{cyan!10}\accCI{45.63}{0.75} 
& \cellcolor{cyan!10}\accCI{24.49}{0.49} 
& \cellcolor{cyan!10}\accCI{85.91}{0.50} 
& \cellcolor{cyan!10}\accCI{93.75}{0.55} 
& \cellcolor{cyan!10}62.44 \\

& MaPLe~\cite{khattak2023maple}
& CVPR-23 & ViT/CLIP
& \accCI{46.72}{0.77}
& \accCI{22.29}{0.45}
& \accCI{88.82}{0.46}
& \accCI{93.15}{0.53}
& 62.75 \\ 

& \cellcolor{cyan!10}\textbf{MaPLe + OURS}
& \cellcolor{cyan!10}-
& \cellcolor{cyan!10}ViT/CLIP
& \cellcolor{cyan!10}\accCI{53.11}{0.86} 
& \cellcolor{cyan!10}\accCI{24.86}{0.51} 
& \cellcolor{cyan!10}\accCI{92.50}{0.39} 
& \cellcolor{cyan!10}\accCI{94.85}{0.46} 
& \cellcolor{cyan!10}66.33 \\  

& LoRA-CLIP~\cite{zanella2024low} 
& CVPR-24 & ViT/CLIP 
& \accCI{50.68}{0.86} 
& \accCI{24.44}{0.51} 
& \accCI{92.63}{0.45} 
& \accCI{96.20}{0.42} 
& 65.99 \\  

& \cellcolor{cyan!10}\textbf{LoRA-CLIP + OURS} 
& \cellcolor{cyan!10}- 
& \cellcolor{cyan!10}ViT/CLIP 
& \cellcolor{cyan!10}\accCI{\textbf{55.95}}{0.77} 
& \cellcolor{cyan!10}\accCI{25.79}{0.50} 
& \cellcolor{cyan!10}\accCI{\textbf{93.43}}{0.41} 
& \cellcolor{cyan!10}\accCI{\textbf{96.88}}{0.45} 
& \cellcolor{cyan!10}\textbf{68.01} \\ 

\hline
\end{tabular}
\end{threeparttable}
\end{adjustbox}
\vspace{-0.4cm}
\end{table*}

Finally, the rectified CLS token feature  at the current 
layer, $f_i^{cls}$, is obtained by multiplying the $Attention_{i}^{cls^{'}}  \in \mathbb{R}^{1 \times (1+n)}   $ with the Value-transformed visual tokens:
\begin{equation}
    f_i^{cls} = Attention_{i}^{cls^{'}} \cdot (v \cdot W_v)  \in \mathbb{R}^{d_v} 
\end{equation}

\subsection{Balanced Alignment and Separation} 
As observed in Fig.~\ref{fig:align+ch} and   Fig.~\ref{fig:loss} in Appendix, while both  modality alignment and class separation improve during training, their simultaneous optimization can lead to an imbalanced learning  of either side. 
To further enhance both criteria, we propose a strategy that balances these two critical objectives. 

Specifically, existing methods adopt only a one-way image-to-text contrastive loss.  
As an  initial  enhancement, we first incorporate the complete bidirectional contrastive loss as defined  in Eqs.~\eqref{eq:clip_full}, \eqref{eq:i2t}, \eqref{eq:t2i}, which serves as our primary loss term: 
\begin{equation}
    L_1 = L_{clip}
\end{equation}

To directly constrain modality alignment, we introduce a secondary loss term, 
$ L_2 $, formulated as the harmonic mean of  \textit{align\_score} and 
\textit{modality\_gap}: 
\begin{equation}
    L_2 = \frac{2\times (1-align\_score)\times modality\_gap}{(1-align\_score) + modality\_gap}
\label{tiaohe}
\end{equation}

Subsequently, we propose a balanced optimization strategy --- Balanced Alignment and Separation (BAS). Unlike static weighting, BAS adaptively adjusts  the contribution of each loss term based on their real-time batch-wise values. 
The  adjustment is achieved through a dynamic weighting coefficient, 
which is determined by the current value of $L_2$:
\begin{equation}
  \beta = \frac{w}{1+e^{-k(L_{2}-T)}}, \quad L_{BAS} = L_{1} + \beta \cdot L_{2}
\label{eq:beta}
\end{equation}
Here, $\beta$ is the sigmoid-shaped weighting coefficient. $w$ sets the maximum initial value of $\beta$, $k$ controls the steepness of the decay, and $T$ defines the threshold of $L_2$ at which the decay becomes significant. When $L_2$ is high (indicating poor modality alignment), $\beta$ approaches $w$, giving more weight to $L_2$ and emphasizing modality alignment. As $L_2$ decreases and falls below the threshold $T$, $\beta$ sharply drops, effectively reducing the influence of $L_2$ and prompting the model to primarily focus on class separation (driven by $L_1$).

During inference,  the image is assigned to the class whose text embedding yields 
the highest cosine similarity with the visual feature.

\section{Experiments}
\subsection{Dataset  and  Implementation Details}
We fine-tune and evaluate CDFSL performance on four target-domain datasets from the BSCD-FSL benchmark~\cite{guo2020broader}: CropDiseases~\cite{mohanty2016using}, EuroSAT~\cite{helber2019eurosat}, ISIC2018~\cite{codella2019skin}, and ChestX~\cite{wang2017chestx}.
We adopt the ViT-Base/16 network as the backbone for all 
experiments. 
Fine-tuning was performed for a total of 100 epochs. Evaluation was consistently conducted 400 times under both the  5-way 1-shot and  5-way 5-shot settings, with the mean classification accuracy reported for all results.
The parameter $\alpha$ in Eq.~\eqref{eqq:alpha} and the parameters $w$, $k$ and $T$ in Eq. ~\eqref{eq:beta} are set to $0.8$ and $7$, $5$, $3.5$, respectively, according to the sensitivity analysis in Appendix~\ref{chaocan}.
All experiments were conducted on a single NVIDIA RTX 4090 GPU.  
For a fair comparison, all baseline methods and our method use the same fixed class-wise textual descriptions. Further details of the text-generation procedure and LLM prompts are provided in Section~\ref{prompt}.

\subsection{Comparison with State-of-the-Art Methods}
For the SF-CDFSL~\cite{DBLP:journals/corr/abs-2411-10070} benchmark setting, we reproduce multiple fine-tuning methods tailored for CLIP, following the experimental protocols exactly as specified in their original studies.  
These competing approaches span representative technical paradigms in PEFT of vision-language models, including prompt learning (CoOp) in  the text
 branch, adapter-based methods (CLIP-Adapter, Tip-Adapter, LoRA-CLIP), 
 and   multi-modal prompt learning (MaPLe)
 —as well as recent advanced tuning strategies (AMU-Tuning, LP++, LDC, StepSPT). 
As presented in Tab.~\ref{tab:main_results},  
our  Semantic Probe framework  consistently enhances the performance of all baseline methods.

\subsection{Ablation Study}
 \textbf{The effects of each component.} 
Tabs.~\ref{tab:ablation} and~\ref{tab:ablation_nolp} report the ablation results with and without LP, respectively. Within each setting, all baselines and their corresponding variants use identical textual information: the LP-enabled methods use the same fixed class-wise descriptions generated by the LLM, whereas the methods without LP use the same class-name prompts. As shown in Tab.~\ref{tab:ablation}, LP alone improves the average accuracy of MaPLe, MMRL, and LoRA-CLIP from 62.75\%, 54.63\%, and 65.99\% to 63.81\%, 55.29\%, and 67.10\%, respectively. With LP fixed, the complete EAR and BAS configuration further improves their average accuracies to 66.33\%, 56.14\%, and 68.01\%, respectively. Tab.~\ref{tab:ablation_nolp} removes LP to isolate the module contributions. Under MaPLe, BAS and EAR individually improve the average accuracy from 62.75\% to 63.96\% and 63.88\%, respectively, while their combination reaches 64.91\%. Under LoRA-CLIP, the corresponding results are 66.50\%, 66.43\%, and 66.85\%, compared with the baseline accuracy of 65.99\%. These results demonstrate that EAR and BAS improve both fine-tuning paradigms independently of LP, with substantially larger gains for MaPLe than for LoRA-CLIP. LoRA-CLIP directly applies low-rank updates to the attention Q/K/V projections and consequently exhibits milder attention collapse. It already provides a strong baseline with limited room for further improvement. Therefore, Semantic Probe primarily aims to revive prompt-based methods that suffer more severe attention collapse under CDFSL.

\begin{table}
\setlength{\abovecaptionskip}{2pt}  
\caption{Ablation study of the 5-way 
 task under MaPLe, MMRL, and LoRA-CLIP (LP denotes using prompts generated by LLM).}
\label{tab:ablation}
\setlength{\tabcolsep}{1.1mm} 
\centering
\resizebox{0.46\textwidth}{!}{      
    \begin{tabular}{ccccccccc}  
        \toprule
        Method  & LP & BAS & EAR & ChestX & ISIC & EuroSAT & CropDiseases & Ave. \\  
        \midrule 
        \multirow{5}{*}{MaPLe} 
        & & & & 22.29 & 46.72 & 88.82 & 93.15 & 62.75 \\  
        & \checkmark & & & 23.62 & 49.02 & 89.58 & 93.00 & 63.81 \\  
        & \checkmark & \checkmark & & 24.10 & 51.70 & 91.77 & 94.13 & 65.43 \\  
        & \checkmark & & \checkmark & 23.70 & 50.96 & 91.99 & 94.58 & 65.31 \\  
        & \checkmark  & \checkmark & \checkmark & \textbf{24.86} & \textbf{53.11} & \textbf{92.50} & \textbf{94.85} & \textbf{66.33} \\  
        \hline 
        \multirow{4}{*}{\begin{tabular}{@{}c@{}}MMRL~\cite{DBLP:conf/cvpr/GuoG25} \\ {\footnotesize (1-shot)}\end{tabular}}
        &  &  &  & 21.95 & 32.99 & 78.39 & \textbf{85.19} & 54.63 \\
        & \checkmark &  &  & 22.18 & 34.52 & 79.76 & 84.72 & 55.29 \\
        & \checkmark &  & \checkmark
        & 22.27 & 35.41 & 80.63 & 84.76 & 55.77 \\
        & \checkmark & \checkmark & \checkmark
        & \textbf{22.36} & \textbf{36.07} & \textbf{81.15}
        & 84.98 & \textbf{56.14} \\
        \midrule
        \multirow{5}{*}{LoRA-CLIP}
        & & & & 24.44 & 50.68 & 92.63 & 96.20 & 65.99 \\  
         & \checkmark & & & 24.86 & 54.55 & 92.78 & 96.22 & 67.10 \\  
        & \checkmark  & \checkmark & & 25.21 & 55.14 & 93.10 & 96.56 & 67.50 \\  
        & \checkmark  & & \checkmark & 25.30 & 55.79 & 93.13 & 96.71 & 67.73 \\  
        & \checkmark  & \checkmark & \checkmark & \textbf{25.79} & \textbf{55.95} & \textbf{93.43} & \textbf{96.88} & \textbf{68.01} \\  
        \bottomrule
    \end{tabular}
}
\vspace{-0.2cm}
\end{table}

\begin{table}
\setlength{\abovecaptionskip}{2pt}
\caption{Ablation study of the 5-way 5-shot task under MaPLe and LoRA-CLIP without LP.}
\label{tab:ablation_nolp}
\setlength{\tabcolsep}{1.1mm}
\centering
\resizebox{0.46\textwidth}{!}{
    \begin{tabular}{cccccccc}
        \toprule
        Method & BAS & EAR & ChestX & ISIC & EuroSAT & CropDiseases & Ave. \\
        \midrule
        \multirow{4}{*}{MaPLe}
        & & & 22.29 & 46.72 & 88.82 & 93.15 & 62.75 \\
        & \checkmark & & 22.95 & 47.20 & 91.83 & 93.84 & 63.96 \\
        & & \checkmark & 22.56 & 46.78 & 91.56 & 94.63 & 63.88 \\
        & \checkmark & \checkmark
        & \textbf{23.51} & \textbf{48.39} & \textbf{92.41} & \textbf{95.31} & \textbf{64.91} \\
        \hline

        \multirow{4}{*}{LoRA-CLIP}
        & & & 24.44 & 50.68 & 92.63 & 96.20 & 65.99 \\
        & \checkmark & & 24.79 & 51.92 & 92.75 & 96.54 & 66.50 \\
        & & \checkmark & 24.88 & 51.27 & 92.96 & 96.62 & 66.43 \\
        & \checkmark & \checkmark
        & \textbf{25.17} & \textbf{52.08} & \textbf{93.28} & \textbf{96.86} & \textbf{66.85} \\ 
        \bottomrule
    \end{tabular}
}
\vspace{-0.5cm}
\end{table}


\textbf{A vanilla strategy of $\beta$ being a selected constant value.} 
As a comparative study, we evaluate a vanilla strategy where the weighting coefficient $\beta$ is a fixed constant throughout training, as opposed to our dynamic BAS approach.  
Tab.~\ref{tab:beta} presents the results for the 5-way 1-shot classification task (400 episodes) on LoRA-CLIP + Ours. We observe that fixed $\beta$ values consistently degrade performance compared to the baseline without our modules (i.e., $\beta=0$), while our dynamic BAS strategy achieves superior results. 
 This confirms that a static weighting scheme is less effective at managing the intricate trade-off between modality alignment and class separation, which our dynamic loss addresses by adaptively adjusting the optimization focus.

\begin{table}
  \caption{Comparison of different fixed $\beta$ values for the 5-way 1-shot classification task.}
  \label{tab:beta}
  \centering
  \scalebox{0.85}{
    \begin{tabular}{cccccc}
      \toprule
      $\beta$ & ChestX & ISIC & EuroSAT & CropDiseases & Ave. \\
      \midrule
      0 & 22.34 & 39.12 & 82.45 & 84.32 & 57.06 \\
      0.1 & 22.11 & 38.26 & 82.40 & 84.07 & 56.71 \\
      0.2 & 22.19 & 38.08 & 82.15 & 83.74 & 56.54 \\
      0.4 & 22.16 & 37.77 & 81.39 & 83.64 & 56.24 \\
      0.6 & 22.08 & 37.82 & 81.69 & 83.83 & 56.36 \\
      0.8 & 22.23 & 37.77 & 82.13 & 83.96 & 56.52 \\
      \midrule
      Ours
        & \textbf{23.65} & \textbf{38.77} & \textbf{82.94}
        & \textbf{85.11} & \textbf{57.62} \\
      \bottomrule
    \end{tabular}
  }
  \vspace{-0.3cm}
\end{table}

\textbf{Controlled attention comparison.}
To isolate the effect of semantic attention guidance, we compare four settings: no attention rectification, EOS-guided EAR, noise-based guidance, and uniform guidance. All four settings use the same LLM-generated descriptions (LP), backbone, LoRA configuration, training procedure, and evaluation protocol. BAS is disabled throughout. As shown in Table~\ref{tab:attention_control}, EOS-guided EAR achieves the best performance on all four datasets, whereas noise-based and uniform guidance are consistently less effective, validating the effectiveness of semantic attention rectification. 

\begin{table}[t]
  \centering
  \caption{Comparison of attention interventions in LoRA-CLIP under the 5-way 5-shot protocol. All four conditions used the same LLM-generated descriptions (LP) and the same training and evaluation settings, with BAS disabled throughout.}
  \label{tab:attention_control}
  \setlength{\tabcolsep}{3.5pt}
  \resizebox{0.48\textwidth}{!}{
    \begin{tabular}{lccccc}
      \toprule
      Attention & ChestX & ISIC & EuroSAT & CropDiseases & Average \\
      \midrule
      None    & 24.86 & 54.55 & 92.78 & 96.22 & 67.10 \\
      Noise    & 24.67 & 49.13 & 90.20 & 90.63 & 63.66 \\
      Uniform  & 24.61 & 49.24 & 91.53 & 96.56 & 65.49 \\
    EOS EAR & \textbf{25.30} & \textbf{55.79} & \textbf{93.13} & \textbf{96.71} & \textbf{67.73} \\
      \bottomrule
    \end{tabular}
  }
\end{table}

\begin{figure}
    \centering
    \includegraphics[width=0.98\linewidth] {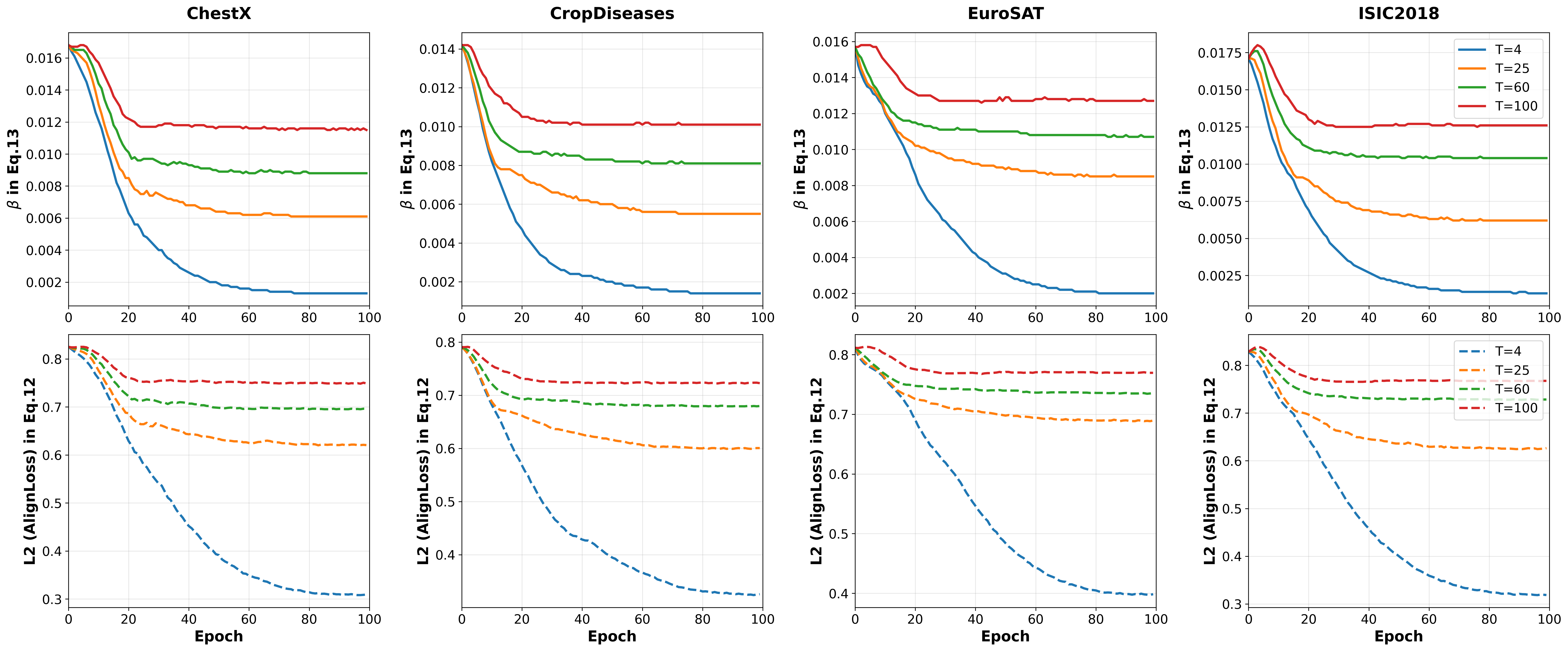}
    \caption{Evolution of the dynamic coefficient $\beta$ in Eq.~(\ref{eq:beta})
    (top) and the alignment loss $L_2$ in Eq.~(\ref{tiaohe}) (bottom) under
    different values of $T$ on the four datasets.}
    \label{fig:beta_l2_temperature}
\vspace{-0.15cm}
\end{figure}

\textbf{Alternative alignment aggregation and weighting strategies.} 
The alignment term combines a positive-pair cosine error and a distance between modality means. In the original formulation, these quantities are combined without additional range normalization. We use the harmonic mean as a symmetric aggregation that reduces domination by a single large component; it is a heuristic choice rather than a theoretically unique one. The dynamic coefficient $\beta$ is detached before it multiplies the alignment loss, so it rescales the alignment gradient without receiving gradients through $\beta(L_2)$. As shown in Fig.~\ref{fig:beta_l2_temperature}, $L_2$ progressively decreases during training, while $\beta$ is dynamically adjusted according to the alignment state. The temperature $T$ controls the sensitivity of this adjustment: smaller values produce sharper reweighting, whereas larger values result in smoother changes. Similar trajectories are observed across all four datasets. As complementary controls, we evaluate normalized harmonic and geometric aggregation, normalized weighted sums with several mixing coefficients, and adaptive sigmoid-, EMA-ratio-, and gradient-ratio weighting strategies in Tab.~\ref{tab:alignment_weight_controls}. Performance remains within a narrow range across these alternatives, indicating that the conclusion is not tied to a single normalized aggregation or adaptive weighting rule.

\begin{table}
\centering
\caption{Comparison of alternative alignment aggregation and weighting strategies for the 5-way 5-shot classification task.}
\label{tab:alignment_weight_controls}
\setlength{\tabcolsep}{10pt}
\renewcommand{\arraystretch}{1.15}
  \resizebox{0.49\textwidth}{!}{
\begin{tabular}{lccccc}
\toprule
\textbf{Loss and weight} & \textbf{ChestX} & \textbf{ISIC} & \textbf{EuroSAT} & \textbf{CropDiseases} & \textbf{Average} \\
\midrule
Raw harmonic & 24.64 & 49.13 & 91.54 & 96.57 & 65.47 \\
Norm harmonic & 24.35 & 49.67 & 91.60 & 96.50 & 65.53 \\
Norm geometric & 24.48 & 49.73 & 91.65 & 96.48 & 65.58 \\
Norm sum, $m=0.25$ & 24.33 & 49.40 & 91.24 & 96.46 & 65.36 \\
Norm sum, $m=0.5$ & 24.50 & 49.41 & 91.39 & 96.50 & 65.45 \\
Norm sum, $m=0.75$ & 24.34 & 49.28 & 91.46 & 96.53 & 65.40 \\
Norm sum, sigmoid & 24.47 & 49.64 & 91.50 & 96.48 & 65.52 \\
Norm sum, EMA & \textbf{24.84} & \textbf{50.49} & \textbf{91.73} & 96.50 & \textbf{65.89} \\
Norm sum, grad & 24.74 & 50.18 & 91.71 & \textbf{96.57} & 65.80 \\
\bottomrule
\end{tabular}
}
\end{table}

\section{Related Work}
\textbf{Cross-Domain Few-Shot Learning }(CDFSL) aims to adapt pretrained  models to distant 
target domains 
(e.g., ISIC2018)  
using limited samples  for rapid generalization. Existing methods can be broadly categorized 
into two types: meta-learning-based approaches \cite{wang2021cross,zhou2023revisiting}, which 
utilize samples in each episode to enable the model to rapidly adapt to the feature 
distribution of the target domain; transfer-based methods 
\cite{liang2021boosting, zou2024attention},
which concentrate solely on enhancing the generalization capability of the model during the training phase. 
Recent studies explore few-shot hyperspectral classification through semantic-anchor-enhanced prototype learning~\cite{YAN2026115822} and orthogonal feature disentanglement~\cite{ZHANG2026114071}. 
We focus on the source-free cross-domain few-shot learning  (SF-CDFSL) setting where only the 
pre-trained model and few-shot target-domain data are accessible. 


\textbf{Parameter-Efficient Fine-Tuning (PEFT)  for Vision-Language Models.} 
Current PEFT strategies for VLMs primarily consist of two paradigms: prompt-based methods~\cite{zhou2022learning, khattak2023maple} that inject trainable vectors into text or visual inputs of frozen encoders; and adapter-based methods that inserted bottleneck modules~\cite{Tip-Adapter, gao2024clip} or low-rank matrix~\cite{zanella2024low}
to efficiently update model parameters. 
 Despite their success for in-domain fine-tuning, the potential and specific tailoring of these methods for VLM-based SF-CDFSL~\cite{Zhao_2026_CVPR} tasks remain underexplored.

\section{Limitations and Future Work}
All experiments in this study use 2D images. To support future extensions, the implementation provides an input adapter for video and volumetric tensors, with explicit temporal or depth dimensions. Mean aggregation or central-frame/slice selection maps these inputs to the current 2D encoder interface. This adapter does not model temporal dynamics or volumetric structure; evaluating dedicated video and 3D extensions remains future work.

\section{Conclusion}
In this work, we establish multiple CLIP-based baselines and identify that CLIP suffers from attention collapse in CDFSL. To address this, we introduce a Semantic Probe framework, comprising an EOS-guided Attention Rectification module and a  dynamic Balanced Alignment and Separation loss, which 
 collaboratively 
improve both modality alignment and class separation. 
Our approach achieves superior mean accuracy across the four target datasets compared with their respective baselines.

\appendix
\section{Detailed Dataset Description}
\begin{figure}
\centering
\includegraphics[width=0.95\linewidth]{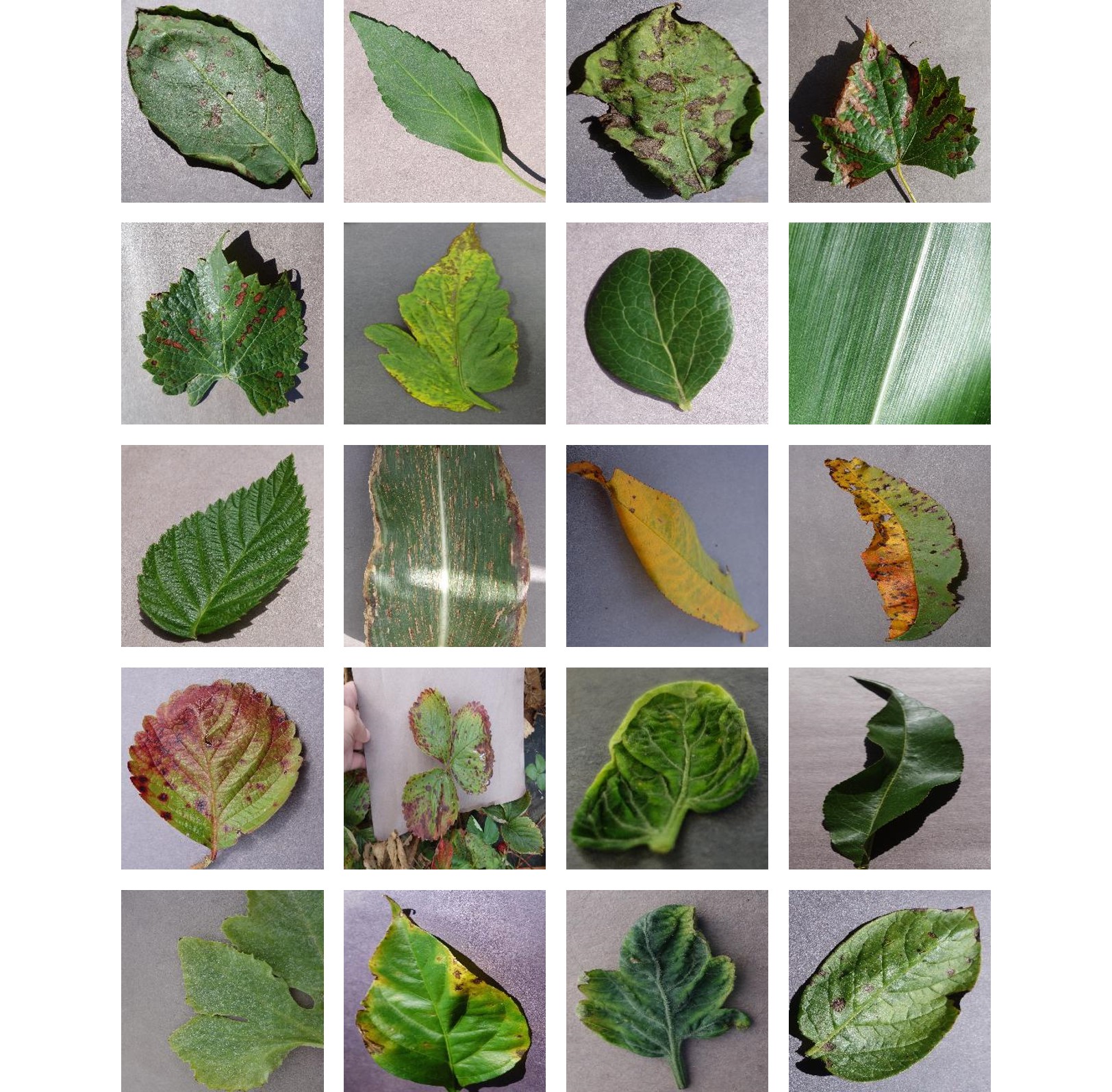} \vspace{-0.2cm}
\caption{Samples from CropDiseases.}
\label{fig:crop}
\vspace{-0.3cm}
\end{figure}

Our experimental setup follows the BSCD-FSL~\cite{guo2020broader} benchmark, addressing the challenge of significant distributional shifts across four distinct target domain datasets. Detailed information on these datasets is provided below:

\textbf{CropDiseases}~\cite{mohanty2016using} is a dataset including 54,306 images of 14 crop species (Apple, Blueberry, Cherry, Corn, Grape, Orange, Peach, Bell Pepper, Potato, Raspberry, Soybean, Squash, Strawberry, and Tomato) with 26 diseases (or healthy). The samples of this dataset are listed in Fig.~\ref{fig:crop}. CropDiseases images are natural images, but are very specialized 
  (specific to the agriculture industry), so the domain gap here is larger than in the previous cross-domain setting~\cite{tseng2020cross}.

\begin{figure}
\centering
\includegraphics[width=0.95\linewidth]{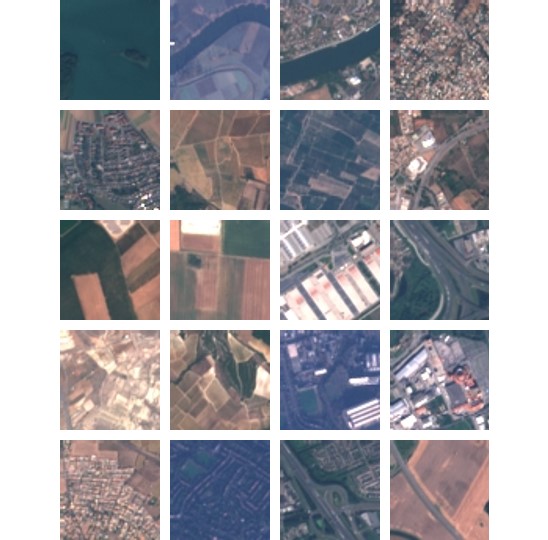} \vspace{-0.2cm}
\caption{Samples from EuroSAT.}
\label{fig:eurosat}
\end{figure}

\textbf{EuroSAT}~\cite{helber2019eurosat} is a dataset for land use and land cover classification. EuroSAT based on Sentinel-2 satellite imagery, covers 13 spectral bands and consists of 10 categories including Industrial Buildings, Residential Buildings, Annual Crop, Permanent Crop, River, Sea \& Lake, Herbaceous Vegetation, Highway, Pasture and Forest, with a total of 27,000 annotated and geographically referenced images. Compared to CropDiseases, EuroSAT images are less similar to \textit{mini}Imagenet (source domain dataset) as they have lost perspective distortion, but are still color images of natural scenes. 
   The samples of this dataset are listed in Fig.~\ref{fig:eurosat}.

\begin{figure}
\centering
\includegraphics[width=0.95\linewidth]{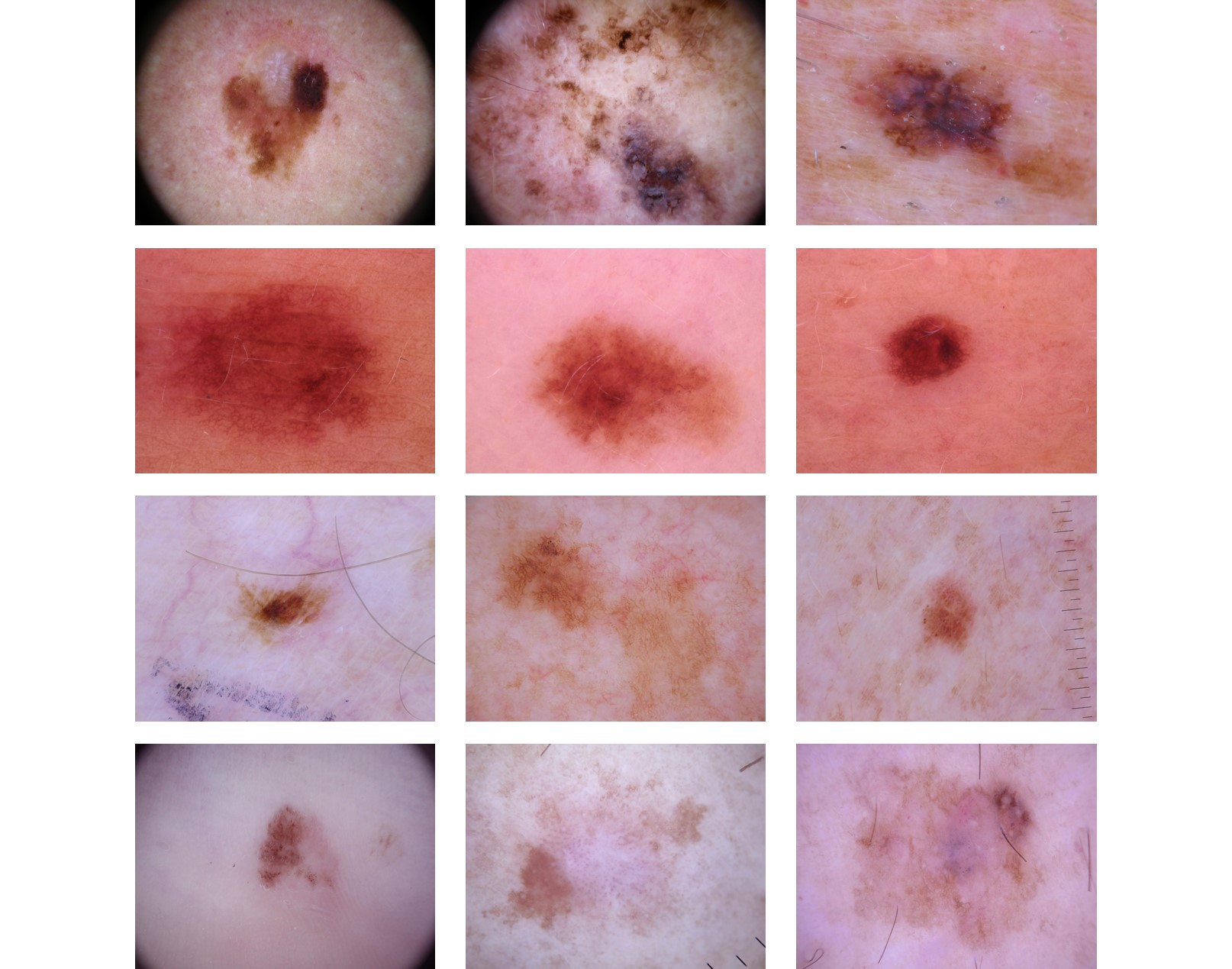} \vspace{-0.2cm}
\caption{Samples from ISIC2018.}
\label{fig:isic}
\end{figure}

The  \textbf{ISIC2018} \cite{codella2019skin} dataset was published by the International Skin Imaging Collaboration (ISIC) as a large-scale dataset of dermoscopy images containing
10,015 images of seven skin injury types (melanoma, melanocytic nevus, basal cell carcinoma, actinic keratosis, benign keratosis, dermatofibroma, or a vascular lesion). ISIC2018 images are even less similar to \textit{mini}Imagenet as they have lost perspective distortion and no longer represent natural scenes.
The samples of this dataset are listed in Fig. \ref{fig:isic}.

\begin{figure}
\centering
\includegraphics[width=0.95\linewidth]{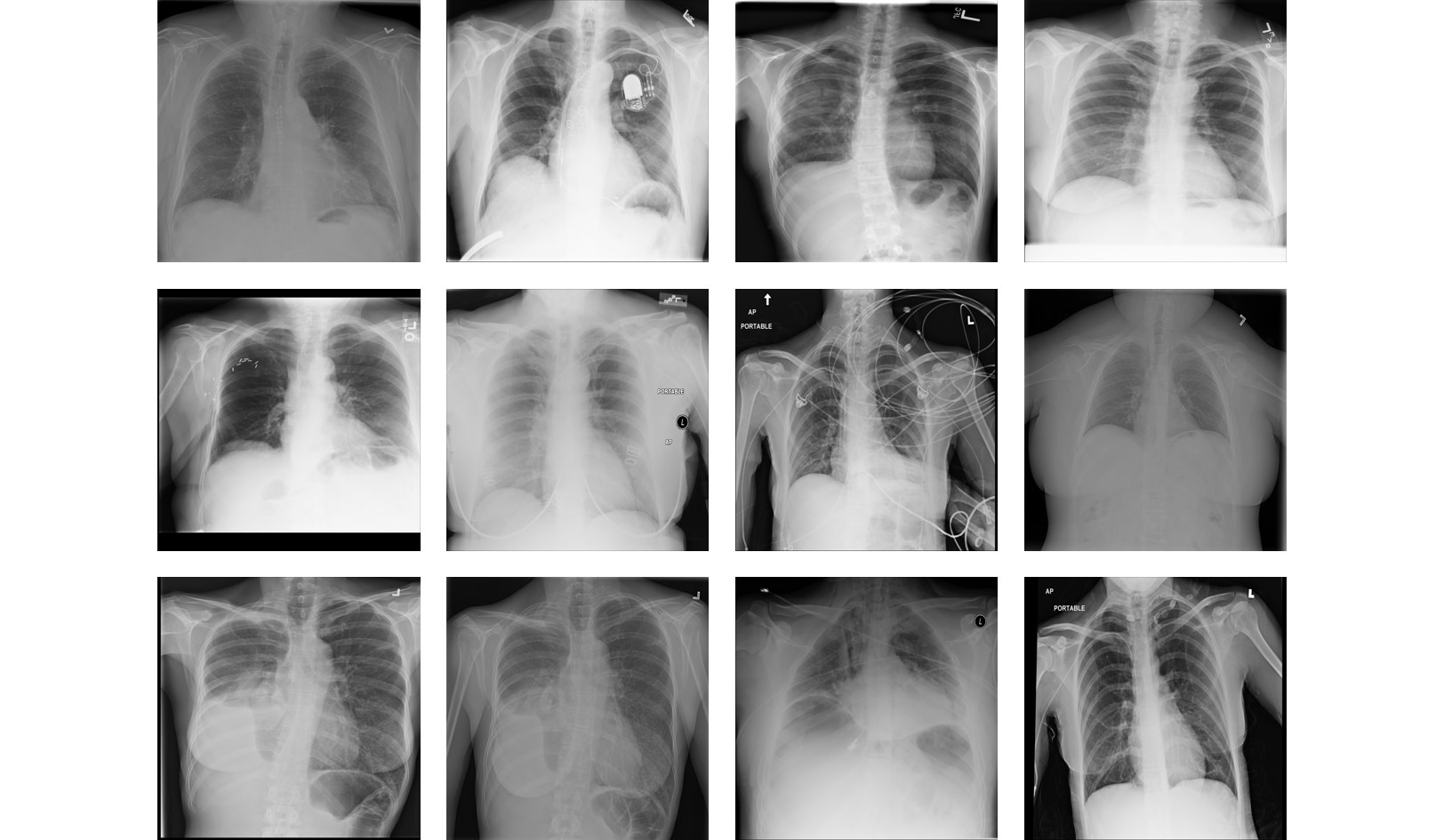} \vspace{-0.2cm}
\caption{Samples from ChestX.}
\label{fig:chestx}
\end{figure}

ChestX-ray14 is the largest lung X-ray database to date, which contains more than 100,000 pre-X-ray views for 14 lung diseases. Categories 1 to 14 correspond to 14 lung diseases, and category 15 indicates no disease. 
\textcite{wang2017chestx} 
 studied the images of eight diseases in this database and constructed the ChestX-ray8 dataset, which comprises 108,948 frontal view X-ray images of 32,717 unique patients with the text-mined eight disease image labels (where each image can have multi-labels) from the associated radiological reports using natural language processing. In this work, we use   \textbf{ChestX-ray8} for cross-domain testing, consistent with 
 \textcite{guo2020broader}.
 ChestX is the most dissimilar to \textit{mini}Imagenet across the four target domains as 
its images 
have lost perspective distortion, do not represent natural scenes, and have lost 2 color channels.
The samples of this dataset are listed in Fig.~\ref{fig:chestx}.

\begin{figure} 
  \centering
  \includegraphics[width=0.95\linewidth]{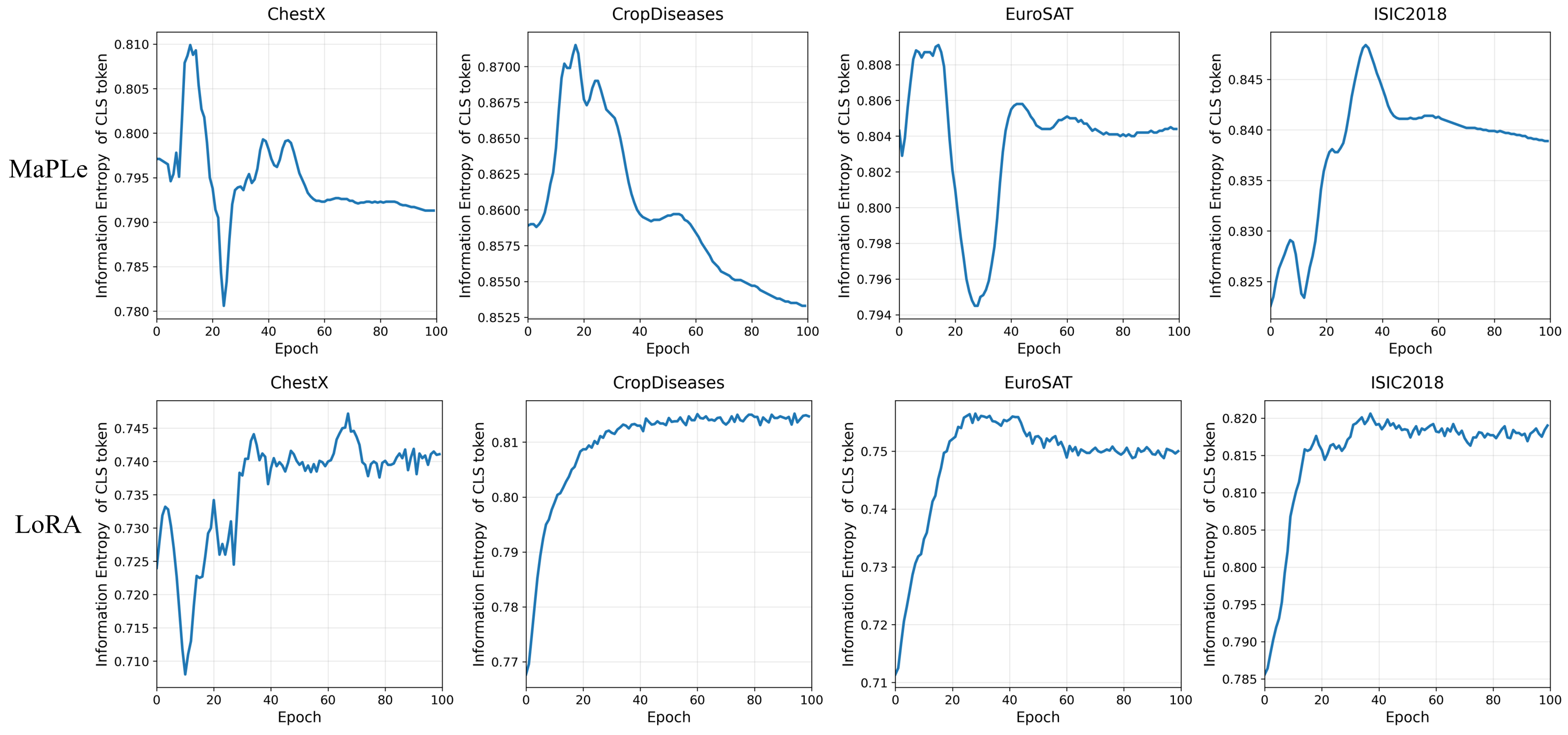}
  \caption{Information entropy of the CLS token during fine-tuning under
  MaPLe (top) and LoRA (bottom).}
  \label{fig:cls-entropy}
\end{figure} 
\section{Quantitative proof of mitigating attention collapse} 
Attention collapse manifests as the visual [CLS] token attending predominantly to itself. Normalized information entropy measures how dispersed CLS attention is over image patches, and thus serves as a direct diagnostic of collapse. As shown in Fig.~\ref{fig:cls-entropy}, the CLS entropy under LoRA rises steadily and stabilizes at a high level during fine-tuning, whereas under MaPLe it remains volatile and decays after an early peak: LoRA continuously disperses the CLS attention, whereas MaPLe fails to sustain this rectification. We therefore report this entropy in Table \ref{tab:token_en} for the raw visual [CLS] token, the textual [EOS] token, and the EAR-rectified [CLS] token. 
Rectification raises the average entropy to 0.8024, indicating more evenly distributed attention with weaker self-focus and thus mitigating cross-domain attention collapse.

Entropy, however, measures spatial dispersion only, not semantic relevance. We therefore complement it with matched patch-deletion interventions using the EAR-rectified CLS attention at a masking fraction of 0.30. Specifically, targeted deletion masks the patches receiving the highest attention weights under EAR, while random deletion masks an equal number of randomly selected patches. As Table~\ref{tab:patch_masking} reports, deleting the patches most attended to by EAR degrades accuracy far more than random deletion on all four datasets. These results demonstrate that the high-attention regions identified by EAR contain prediction-relevant evidence, confirming that the EAR-rectified attention is semantically informative rather than merely more dispersed.

\begin{table}[thbp]
  \centering
  \caption{Information Entropy of token attention over image patches.} 
  \label{tab:token_en}
  \scalebox{0.7}{ 
  \begin{tabular}{lccc}
    \toprule
    Dataset & Visual [CLS]  token & Textual [EOS]  token & Rectified [CLS]  token \\
    \midrule
    Ches. & 0.6100 &  0.7887 & 0.7094 \\
    ISIC. & 0.7646 &  0.8758 & 0.8344 \\
    Euro. & 0.8182 &  0.8593 & 0.8604 \\
    Crop. & 0.7284 &  0.8991 & 0.8055 \\
    \midrule
    Ave.  & 0.7303 &  0.8557 & 0.8024 \\
    \bottomrule
  \end{tabular}
  }
\end{table}

\begin{table}[thbp]
  \centering
  \caption{Query patch masking results at a masking fraction of 0.30, reported as accuracy (\%) averaged over 400 episodes.}
  \label{tab:patch_masking}
  \scalebox{0.8}{
  \begin{tabular}{lccc}
    \toprule
    Dataset & Unmasked & Top patch deletion & Random deletion \\
    \midrule
    ChestX & 24.39 & 21.07 & 22.03 \\
    ISIC & 49.48 & 32.32 & 41.28 \\
    EuroSAT & 91.45 & 64.81 & 72.02 \\
    CropDiseases & 96.48 & 71.66 & 86.44 \\
    \bottomrule
  \end{tabular}
  }
\end{table}

\begin{figure*}[thbp]
    \centering
    \includegraphics[width=\textwidth]{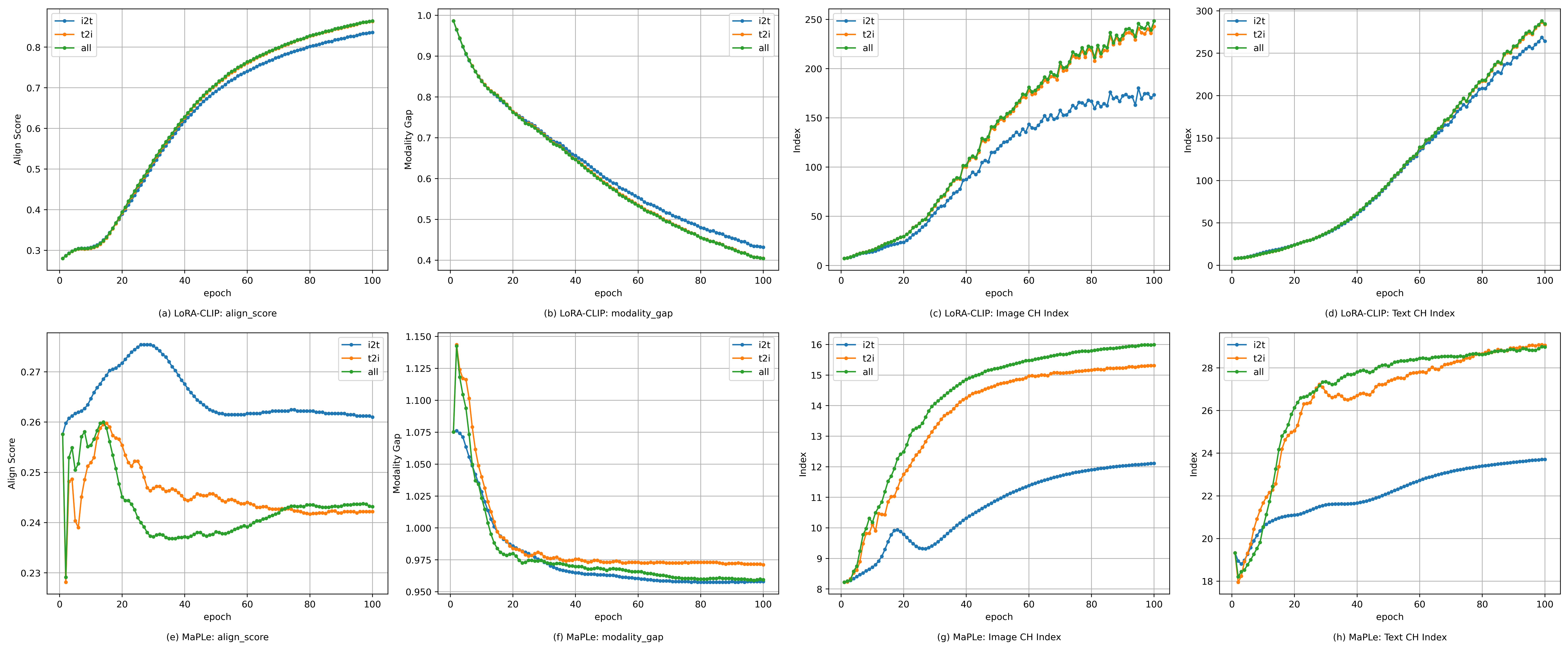}   
    \caption{
	The alignment/aggregation metrics under different losses across training epochs. The modality alignment is only marginally affected by the choice of loss function, whereas the class aggregation improves significantly after the text-to-image term $L_{t \to i}$ is introduced. This suggests that CLIP’s InfoNCE loss primarily emphasizes class 
    separation, 
    with relatively weak constraints on modality alignment.
    }
    \label{fig:loss}
    \vspace{-0.3cm}
\end{figure*}

\section{Contrastive loss lacks constrain to modality  alignment but rather class aggregation}
\label{author info}
To analyse the underlying reasons for the performance gap between LoRA-CLIP and MaPLe, 
we 
 also 
examine the role of the original loss function of CLIP. CLIP  uses InfoNCE 
\cite{DBLP:journals/corr/abs-1807-03748} loss to train the model, which consists of two 
parts:
\begin{itemize}
    \item Image-to-text contrastive loss: 
    \begin{equation}
        L_{i \to t} = - \frac{1}{B} \sum_{i=1}^B \log \frac{\exp(\text{sim}(v_i, t_i)/\tau)}{\sum_{j=1}^{B} \exp(\text{sim}(v_i, t_j)/\tau)} 
    \label{eq:i2t}
    \end{equation}
    \item Text-to-image contrastive loss: 
    \begin{equation}
        L_{t \to i} =  - \frac{1}{B} \sum_{i=1}^B \log \frac{\exp(\text{sim}(t_i, v_i)/\tau)}{\sum_{j=1}^{B} \exp(\text{sim}(t_i, v_j)/\tau)}
     \label{eq:t2i}
    \end{equation}
\end{itemize} 
where $\text{sim}(a, b) = \frac{a^\top b}{\|a\|\|b\|}$ represents cosine similarity, and 
$\tau$ is a temperature parameter.

Existing CDFSL baselines rely solely on the image-to-text contrastive loss term $L_{i  \to t}$ for optimization. 
Some works like 
\textcite{liang2022mind} and \textcite{tyshchuk2023isotropy} 
argue that CLIP's contrastive loss has minimal impact on guiding modality alignment.
This view is supported by Figure \ref{fig:loss}, which compares the modality alignment metrics of LoRA-CLIP (panels a, b) and MaPLe (panels e, f) under different contrastive loss settings. Notably, when the text-to-image contrastive loss $L_{t  \to i}$
  is incorporated alongside $L_{i  \to t}$, the  resulting  curves have no obvious changes. This empirical evidence strongly indicates that \textbf{the  text-to-image contrastive loss has a limited ability to constrain modality alignment.}

Fig.\ref{fig:loss} (c, d) and (g, h) also present the changes in the CH index under different contrastive losses for LoRA-CLIP and MaPLe, respectively.
 It can be observed that the CH index of both visual and textual features is improved after using $L_{t \to i}$, and it can be further enhanced by using $L_{i  \to t} + L_{t  \to i}$. This suggests  that \textbf{the role of the InfoNCE loss in CLIP is to strengthen the feature aggregation  within the same class.}

\section{Prompts for LLMs and Responses}
\label{prompt}
\subsection{Prompts for LLMs} 
To align with CLIP's pre-training paradigm and enrich textual modality information~\cite{schrodi2024two},
we  employ Llama3-8B-Instruct~\cite{grattafiori2024llama} with 5 different LLM prompts to generate 10 class-specific descriptions per prompt, 
yielding a total of 50 semantically rich text descriptions per class. During the fine-tuning phase of each episode, we randomly select 30 descriptions per support set class as textual prompts. The class-wise mean of the resulting text features is computed and used as weights for the text classifier. 
The five distinct LLM prompts are as follows:
\begin{enumerate}
\item Describe what \{article\} \{category\} (a kind of \{domain\_word\}) looks like in \{max\_len\} words. Do not include any extra words or phrases.

\item How can you identify \{article\} \{category\}  (a kind of \{domain\_word\})? In \{max\_len\} words.

\item Do not include any extra words or phrases. What are the identifying characteristics of \{article\} \{category\}  (a kind of \{domain\_word\}) in \{max\_len\} words?

\item Do not include any extra words or phrases. What does \{article\} \{category\}  (a kind of \{domain\_word\}) look like? In \{max\_len\} words. Do not include any extra words or phrases.

\item Describe an image of \{article\} \{category\}  (a kind of \{domain\_word\}) in \{max\_len\} words. Do not include any extra words or phrases.
\end{enumerate}

where \{category\} represents the category name, \{article\} represents the article (a/an) chosen based on the category name, \{domain\_word\} is the domain descriptor corresponding to the dataset, and \{max\_len\} is used to control the length of the generated sentence to prevent exceeding the specified maximum token limit (77) of CLIP's text encoder.

 

 The specific domain description words and category names used in the prompts for different datasets are presented in Tab.\ref{tab:datasets} .
\begin{table*}
  \centering
  \caption{Dataset Information}
  \begin{tabular}{ccc}
    \toprule
  \textbf{Dataset} & \textbf{Domain Description Word} & \textbf{(Partial) Category Names} \\
    \midrule
    \multirow{5}{*}{CropDiseases} & \multirow{5}{*}{crop leaf} & Apple\_\_Apple\_scab \\
     &  & Blueberry\_\_healthy \\
     &  & Grape\_\_Black\_rot \\
     &  & Tomato\_\_Bacterial\_spot \\
     &  & Corn\_(maize)\_\_Northern\_Leaf\_Blight \\
    \midrule
    \multirow{4}{*}{EuroSAT} & \multirow{4}{*}{satellite image} & Annual Crop Land, Sea or Lake, Forest, \\
     &  & Herbaceous Vegetation Land, Highway or Road, \\
     &  & Pasture Land, Industrial Buildings, Permanent \\
     &  & Crop Land, River, Residential Buildings \\
    \midrule
    \multirow{3}{*}{ISIC} & \multirow{3}{*}{skin lesion image} & Melanoma, Melanocytic Nevus, Basal, Cell \\
     &  & Carcinoma, Benign Keratosis, Actinic Keratosis, \\
     &  & Dermatofibroma \\
    \midrule
    \multirow{2}{*}{ChestX} & \multirow{2}{*}{diseased lung X-ray} & Atelectasis, Cardiomegaly, Effusion, \\
     &  & Infiltration, Mass, Nodule, Pneumothorax \\
    \bottomrule
  \end{tabular}
  \label{tab:datasets}
\end{table*}


\subsection{Responses} 
We generate fifty descriptions per class with Meta-Llama-3-8B-Instruct using five prompt templates and greedy decoding. At each decoding step, the model selects the token with the highest predicted probability without stochastic token sampling; therefore, no sampling temperature or token-sampling seed is used. All baseline methods and our method use the same fixed class-wise textual descriptions, ensuring that the available textual information is consistent across comparisons. 
Here we show two example responses of LLMs for a specific category within each dataset.

\begin{enumerate}
        \item \textit{Apple\_\_\_Apple\_scab class in  CropDiseases:}
        \begin{itemize}
            \item A leaf with Apple Scab disease is typically yellow or pale green with large, circular, dark green or black lesions that are often surrounded by a yellow or white rim.
            \item Identify apple leaf with apple scab by observing: yellow-green, irregularly shaped spots with darker borders; powdery, white growth on underside; leaves may be distorted, curled, or crinkled; fungal odor; affected leaves often fall prematurely; can occur on both sides of leaf.
        \end{itemize}
        
            \item \textit{Annual Crop Land class in  EuroSAT:}
        \begin{itemize}
            \item A satellite image of annual crop land appears as a mosaic of rectangular fields with varying shades of green, often with narrow strips of darker or lighter tones indicating different crop types, irrigation patterns, and soil conditions.
            \item The image shows a vast expanse of green fields, rectangular in shape, with straight edges and roads connecting them. The fields are divided into smaller sections, some with darker or lighter shades of green, indicating varying crop growth stages.
        \end{itemize}

            \item \textit{Melanoma class in ISIC2018:}
        \begin{itemize}
            \item Irregularly shaped, dark brown or black mole with uneven borders, multiple colors, and diameter greater than 6 millimeters.
            \item Abnormal moles with asymmetry, irregular borders, varied colors, and diameter greater than 6mm.
        \end{itemize}
        
            \item \textit{ Atelectasis class in  ChestX:} 
        \begin{itemize}
            \item X-ray images of atelectasis show a collapsed lung with a dense, opaque appearance and loss of lung markings.
            \item Atelectasis appears as a dark or lucent area in the lung, often with a sharp border and surrounding hyperinflation.
        \end{itemize}

\end{enumerate}

\begin{figure*}
    \centering
    \includegraphics[width=\linewidth]{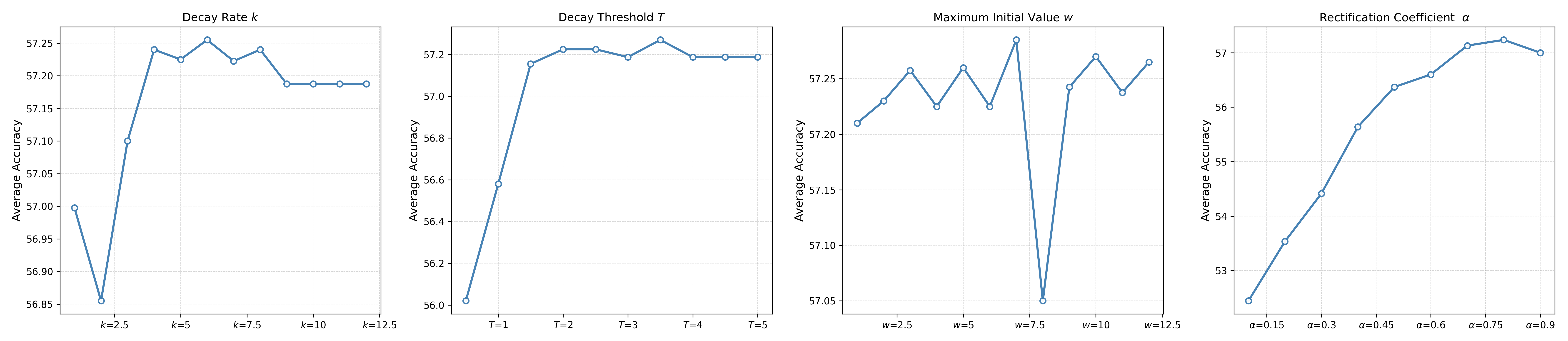} 
    \caption{
	Performance of our Semantic Probe  method under different configurations: (a) the  steepness parameter $k$, (b)  the decay threshold $T$, (c) the 
    maximum initial value $w$, (d) the EOS-guided 
    attention rectification weighting coefficient  $\alpha$.
    Here, $w > 0$ defines the upper bound of $\beta$, $k > 0$ controls the steepness of the sigmoid transition, and $T \geq 0$ is the value of $L_{2}$ at which $\beta = w/2$.
    } %
    \vspace{-0.3cm} 
    \label{fig:4plot}
\end{figure*}

\begin{figure*}
\centering
\subfloat[]{\includegraphics[width=0.47\linewidth]{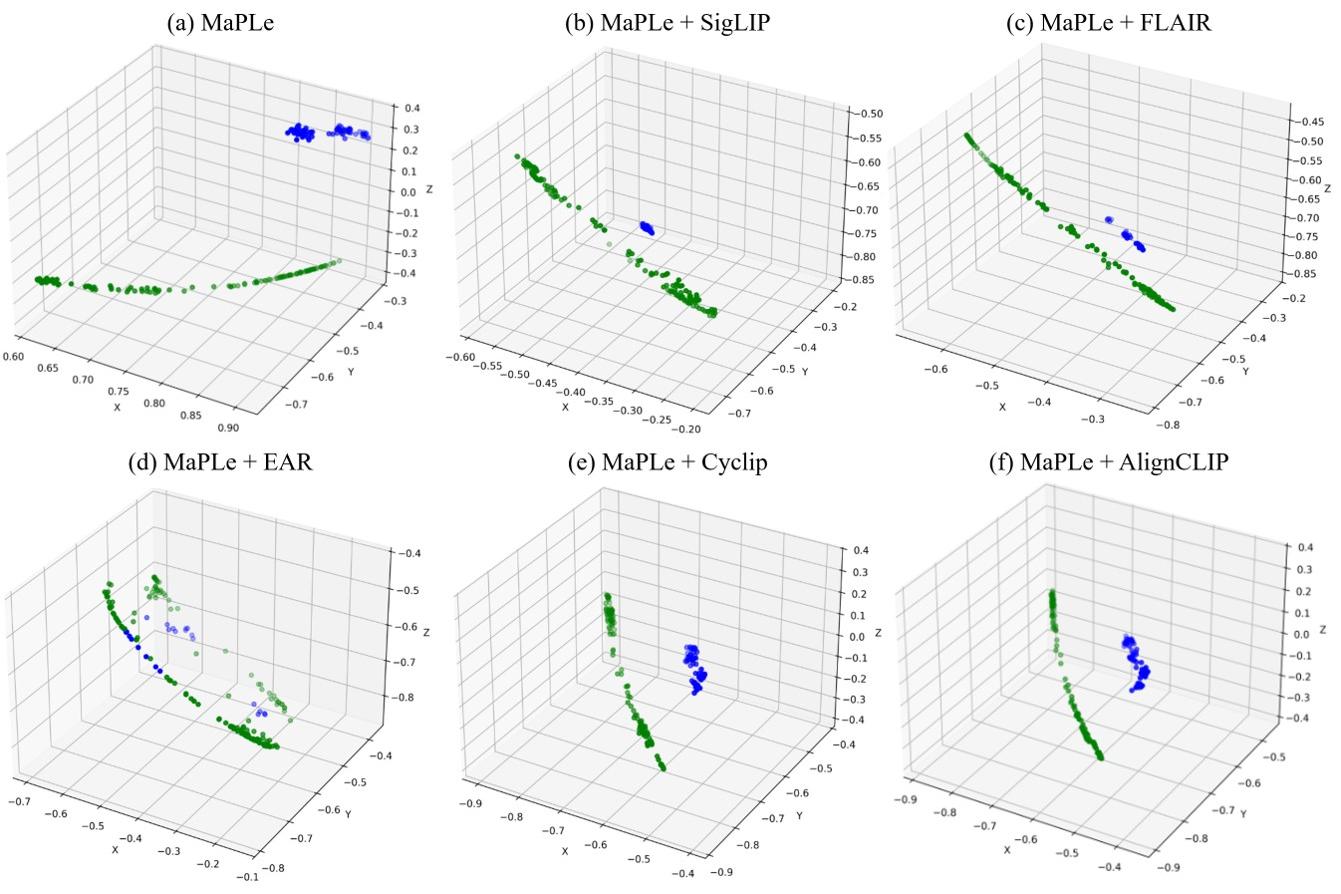}%
\label{fig_first_case}}
\hfil
\subfloat[]{\includegraphics[width=0.5\linewidth]{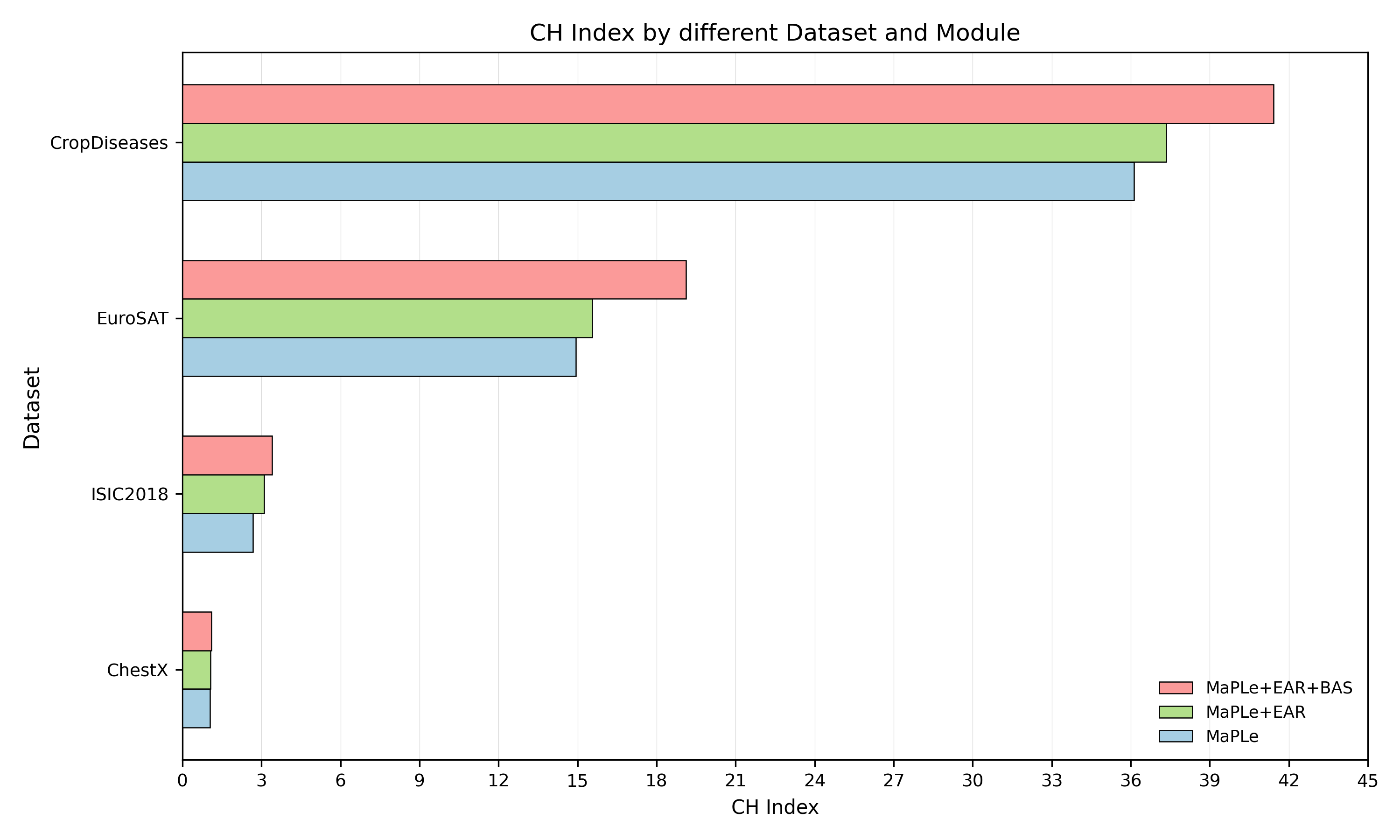}%
\label{fig_second_case}}
\caption
{
(a) DOSNES visualization of features under various CLIP-based methods. (b)  The CH index of visual features across CDFSL target datasets.
}
\label{fig:verfi}
\vspace{-0.3cm}
\end{figure*}

\section{Hyperparameter Analysis}
\label{chaocan}
We conduct a thorough analysis of the key hyperparameters in our 
proposed 
 Semantic Probe framework 
to validate their influence on model performance. The parameters governing the Balanced Alignment and Separation (BAS) loss are the sigmoid steepness $k$, the decay threshold $T$, and the maximum initial value $w$, as defined in Eq.~\eqref{eq:beta}. For the EOS-guided Attention Rectification (EAR) module, the critical hyperparameter is the weighting coefficient $\alpha$. Our analysis is performed on the 5-way 1-shot task, with results averaged across all four target datasets for clarity.

For the steepness parameter $k$, shown in Fig.\ref{fig:4plot}a, the average accuracy increases as $k$ rises, indicating that a sharper transition from modality alignment to class separation is generally beneficial. Performance then plateaus, demonstrating that the model is not overly sensitive to the exact value once an effective level of steepness is achieved. 
Fig.\ref{fig:4plot}b shows the impact of the threshold $T$, which dictates when the transition begins. Performance improves as the threshold increases, reaching its maximum at $T=3.5$. A high threshold causes the model to pivot to class separation before modality alignment is sufficient, whereas a low threshold may delay this transition unnecessarily. The optimal value of $T=3.5$ allows for a balanced training process.
Fig.\ref{fig:4plot}c demonstrates the effect of the maximum initial value $w$. The average accuracy remains relatively stable across the tested range of $w$ values, peaking at $w=7$. This suggests that the model's performance is robust to the initial weighting of the alignment loss, as long as it provides sufficient guidance without dominating the total loss.

The weighting coefficient $\alpha$ (defined in Eq.~\eqref{eqq:alpha}) in the EAR module determines the balance between the original CLS token's attention and the textual EOS token's guidance. As shown in Figure \ref{fig:4plot}d, when $\alpha$ is low (e.g., 0.1), the influence of the original CLS attention is too small, resulting in suboptimal performance. As we increase $\alpha$, performance improves significantly, as the model effectively integrates the guided attention with its own learned patterns. The performance peaks at $\alpha=0.8$, demonstrating that a 
enough  
but not overwhelming influence from the textual modality is most effective.

\section{Layer placement of the EAR module} 
EAR is only applied to the final layers of CLIP. This design choice was motivated by prior work, particularly MMA~\cite{Yang_2024MMA_CVPR}  and MMRL~\cite{DBLP:conf/cvpr/GuoG25}, which established two key findings:
\begin{itemize}
  \item Deep layers encode task-specific discriminative knowledge for fine-tuning; shallow layers preserve universal cross-domain knowledge and should be frozen. 
  \item Textual features are more discriminative than visual features, and the semantic gap between modalities is larger at shallow layers, making cross-modal alignment more challenging.
\end{itemize}

Additional layer experiments (5-way 1-shot) are shown in Table~\ref{tab:layer}. Deep-layer placement yields the best performance across all datasets, consistent with the above analysis. 
\begin{table}
   \caption{Ablation study on different layer groups for the 5-way 1-shot classification task.} 
  \label{tab:layer}
\centering
\begin{tabular}{lccccc}
\toprule
Layers & Ches. & ISIC. & Euro. & Crop. & Ave. \\
\midrule
1-4   & 21.86 & 39.25 & 82.37 & 84.44 & 56.98 \\
5-8   & 21.95 & 39.95 & 82.57 & 84.76 & 57.31 \\
9-12  & 22.34 & 39.96 & 83.24 & 84.82 & 57.39 \\
\bottomrule
\end{tabular}
\end{table}

\section{Implementation with Other Losses} 
Table~\ref{tab:clipsota} compares adaptations of several vision-language loss 
formulations within the MaPLe few-shot fine-tuning framework. The 
SigLIP~\cite{zhai2023sigmoid}-inspired implementation retains sample-level image-text 
Sigmoid contrastive learning, while the FLAIR~\cite{xiao2024flair}-inspired 
implementation preserves fine-grained interactions between textual features and local 
visual features and applies its corresponding Sigmoid contrastive objectives at the 
sample level. For CyCLIP~\cite{goel2022cyclip}, because multiple support images of the 
same class correspond to a class-level textual representation in the cross-domain few-
shot setting, its original intra-batch relational constraints are adapted to operate on 
class-mean representations. For AlignCLIP~\cite{eslami2024mitigate}, textual-feature 
similarities replace the additional text encoder used in its SharedCLIP implementation, 
 and class-mean representations are likewise used for components involving intra-batch 
 image-text relationships. Therefore, class-mean replacement is not shared by all 
 compared losses. These results should be interpreted as implementation-specific loss 
 adaptations within the MaPLe framework rather than direct evaluations of the original 
 pretrained models. 

\begin{table}[thbp] 
\setlength{\abovecaptionskip}{2pt}   
  \caption{Comparison of adapted vision-language losses within the MaPLe fine-tuning framework on the 5-way 5-shot  
 task.}  
  \label{tab:clipsota} 
  \centering  
  \scalebox{0.7}{  
      \begin{tabular}{lccccc}  
        \toprule 
        Method &  ChestX & ISIC & EuroSAT & CropDiseases & Ave. \\ 
        \midrule 
        MaPLe & 23.62 & 49.02 & 89.58 & 93.00 & 63.81 \\ 
        MaPLe + SigLIP & 20.08 & 22.79 & 61.87 & 54.32 & 39.77 \\ 
        MaPLe + FLAIR & 20.07 & 22.77 & 61.84 & 54.30 & 39.75 \\ 
        MaPLe + CyCLIP & 24.40 & 51.58 & 92.06 & 94.19 & 65.56 \\ 
        MaPLe + AlignCLIP & 24.19 & 51.78 & 91.78 & 94.20 & 65.49 \\ 
        MaPLe + \textbf{Ours} & \textbf{24.86} & \textbf{53.11} & \textbf{92.50} & \textbf{94.85} & \textbf{66.33} \\ 
        \bottomrule 
      \end{tabular} 
    } 
     \vspace{-0.3cm} 
\end{table} 

\section{Module Verification}
Fig.\ref{fig:verfi}a   illustrates    the feature distributions of visual and textual modalities for various CLIP-based methods using DOSNES~\cite{lu2016doubly}  visualization. 
Initially, MaPLe exhibits a distinct separation between visual and textual features. 
While methods such as SigLIP, FLAIR, CyCLIP, and AlignCLIP reduce the center distance between modalities, they fail to  effectively achieve feature fusion. In contrast, MaPLe with our EAR  significantly   enhances modality integration.

Fig.\ref{fig:verfi}b reports the  Calinski-Harabasz (CH) index 
for visual features across three configurations: 
MaPLe, MaPLe with EAR, and MaPLe with EAR + BAS. The results indicate that: (1) EAR strengthens class separation within the visual modality; and (2) when combined with BAS, the class separation is further improved due to enhanced modality alignment. 
These findings confirm that improved modality alignment fosters better class separation, a critical factor for superior CDFSL performance.

\section{Comparison with Representative CDFSL Methods} 
Tables \ref{tab:1shot} and \ref{tab:5shot} provide a comprehensive comparison of our proposed 
Semantic Probe framework 
with state-of-the-art approaches, including 
MEM-FS~\cite{DBLP:journals/tip/WalshOS23}, StyleAdv~\cite{DBLP:conf/cvpr/FuXFJ23}, FLoR~\cite{DBLP:conf/cvpr/ZouLH0024}, 
DAMIM~\cite{DBLP:conf/aaai/MaZ0025}, 
AttnTemp~\cite{zou2024attention}, 
CD-CLS~\cite{zou2024a}, 
PMF~\cite{DBLP:conf/cvpr/Hu0SKH22}, 
IM-DCL~\cite{DBLP:journals/tip/XuLZFSCL24}, 
and StepSPT~\cite{DBLP:journals/corr/abs-2411-10070}, 
 for 5-way 1-shot and 5-shot classification tasks. These methods employ 
varied backbones  
and training configurations, differing in whether they utilize source domain pre-training (e.g., miniImageNet) or target domain fine-tuning  (FT). 
LoRA-CLIP with Semantic Probe shows strong performance across the four target datasets, achieving mean accuracies of 57.62\% in the 1-shot setting and 68.01\% in the 5-shot setting.

 It is noteworthy that IM-DCL \cite{DBLP:journals/tip/XuLZFSCL24} achieves superior performance on the ChestX dataset, which we attribute to its ResNet-based backbone being better suited for capturing local features prevalent in medical images.

\begin{table*}[thbp] 
\setlength{\abovecaptionskip}{2pt}  
  \caption{Comparison with 
 representative 
 works on the 5-way 1-shot  classification.}      
  \label{tab:1shot} 
  \centering 
  \scalebox{0.8}{ 
      \begin{tabular}{lccccccccc} 
        \toprule 
        Method & Backbone & Mark & Source & Target & ChestX & ISIC & EuroSAT & CropDiseases & Ave. \\ 
        \midrule 
        MEM-FS & ViT/DINO & TIP-23 & & & 22.76 & 32.97 & 68.11 & 81.11 & 51.24\\
        StyleAdv & ViT/DINO &  CVPR-23  & \checkmark & - & 22.92 & 33.05 & 72.15 & 81.22 & 52.34\\
        FLoR & ViT/DINO & CVPR-24  & \checkmark & - & 22.78 & 34.20 & 72.39 & 81.81 & 52.80 \\
        DAMIM & ViT/DINO & AAAI-25 & \checkmark & - & 22.97 & 34.66 & 72.87 & 82.34 & 53.21 \\
        AttnTemp & ViT/DINO &  NeurIPS-24 & \checkmark & - & 23.19 & 34.92 & 74.35 & 84.02 & 54.12 \\
        CD-CLS & ViT/DINO &  NeurIPS-24 & \checkmark & \checkmark & 23.39 & 35.56 & 74.97 & 84.53 & 54.62 \\
        PMF & ViT/DINO & CVPR-22 & \checkmark & \checkmark & 21.73 & 30.36 & 70.74 & 80.79 & 50.91 \\
        StyleAdv-FT & ViT/DINO &  CVPR-23  & \checkmark & \checkmark & 22.92 & 33.99 & 74.93 & 84.11 & 53.99 \\
        FLoR-FT & ViT/DINO & CVPR-24 & \checkmark & \checkmark & 23.26 & 35.49 & 73.09 & 83.55 & 53.85 \\
        DAMIN-FT & ViT/DINO & AAAI-25 & \checkmark & \checkmark & 23.38 & 36.35 & 73.61 & 83.90 & 54.31 \\
        AttnTemp-FT & ViT/DINO & NeurIPS-24  & \checkmark & \checkmark & 23.63 & 38.05 & 75.09 & 84.78 & 55.39 \\
        IM-DCL& RN10 & TIP-24 & -& \checkmark &\textbf{23.98} &38.13 &77.14 & 84.37& 55.91\\
        StepSPT & ViT/CLIP & TPAMI-25 & - & \checkmark & 22.84 & 32.97& 70.01 & 84.84 & 52.68 \\
        \midrule
        MaPLe  & ViT/CLIP &  CVPR-23 & - & \checkmark & 
        20.89 & 31.10 & 75.66 & 81.87 & 53.04 \\
        \rowcolor{cyan!10}
        MaPLe + Ours & ViT/CLIP & - & - & \checkmark & 
        21.41  & 34.27 & 82.94 & 82.83 & 55.36 \\  
        \textcolor{teal}{\small		\textbf{$\Delta$}} & -	& - & - & - &	\textcolor{teal}{\small	\textbf{+0.52}} & 	\textcolor{teal}{\small	\textbf{+3.17}} & 		\textcolor{teal}{\small		\textbf{+7.28}} & 		\textcolor{teal}{\small		\textbf{+0.96}} & 		\textcolor{teal}{\small		\textbf{+2.32}} \\  \hline
         LoRA-CLIP  & ViT/CLIP &  CVPR-24 & - & \checkmark & 21.73 & 34.45 & 80.34 & 84.95 & 55.37 
         \\ 
        \rowcolor{cyan!10}
        LoRA-CLIP + Ours & ViT/CLIP & - & - & \checkmark & 23.65 & \textbf{38.77} &    \textbf{82.94} & \textbf{85.11} & \textbf{57.62} \\
        \textcolor{teal}{\small		\textbf{$\Delta$}} & -	& - & - & - &	\textcolor{teal}{\small	\textbf{+1.92}} & 	\textcolor{teal}{\small	\textbf{+4.43}} & 		\textcolor{teal}{\small		\textbf{+2.60}} & 		\textcolor{teal}{\small		\textbf{+0.16}} & 		\textcolor{teal}{\small		\textbf{+2.25}} \\  
        \bottomrule
      \end{tabular}
    } 
\end{table*} 
\begin{table*}[H]
\setlength{\abovecaptionskip}{2pt}  
  \caption{Comparison with 
 representative 
 works on the 5-way 5-shot classification.}  
  \label{tab:5shot}
  \centering
  \scalebox{0.8}{ 
      \begin{tabular}{lccccccccc}
        \toprule
        Method & Backbone &  Mark & Source & Target & ChestX & ISIC & EuroSAT & CropDiseases & Ave. \\
        \midrule 
        MEM-FS & ViT/DINO & TIP-23 & & & 26.67 & 47.38 & 86.49 & 93.74 & 63.57\\
        StyleAdv & ViT/DINO &  CVPR-23  &\checkmark & - & 26.97 & 47.73 & 88.57 & 94.85 & 64.53\\
        FLoR & ViT/DINO & CVPR-24  & \checkmark & - & 27.28 & 49.52 & 90.41 & 95.28 & 65.48 \\
        DAMIM & ViT/DINO & AAAI-25 & \checkmark & - & 27.28 & 50.76 & 89.50 & 95.52 & 65.77 \\
        AttnTemp & ViT/DINO &  NeurIPS-24 & \checkmark & - & 27.72 & 53.09 & 90.13 & 95.53 & 66.62 \\
        CD-CLS & ViT/DINO &  NeurIPS-24 & \checkmark & \checkmark & 27.66 & 54.69 & 91.53 & 96.27 & 67.54 \\
        PMF & ViT/DINO & CVPR-22 & \checkmark & \checkmark & 27.27 & 50.12 & 85.98 & 92.96 & 64.08 \\
        StyleAdv-FT & ViT/DINO &  CVPR-23  & \checkmark & \checkmark & 26.97 & 51.23 & 90.12 & 95.99 & 66.08 \\
        FLoR-FT & ViT/DINO & CVPR-24 & \checkmark & \checkmark & 27.02 & 53.06 & 90.75 & 96.47 & 66.83 \\
        DAMIN-FT & ViT/DINO & AAAI-25 & \checkmark & \checkmark & 
        27.82
        & 54.86 & 91.18 & 96.34 & 67.78 \\
        AttnTemp-FT & ViT/DINO & NeurIPS-24  & \checkmark & \checkmark & 28.03 & 54.91 & 90.82 & 96.66 & 67.61 \\
        IM-DCL &  RN10 & TIP-24  & - & \checkmark & \textbf{28.93} & 52.74 & 89.47 & 95.73 & 66.72 \\
        StepSPT & ViT/CLIP & TPAMI-25 &  - & \checkmark & 26.36 & 52.12 & 89.40 & 96.01 & 65.97 \\
        \midrule
        MaPLe  & ViT/CLIP &  CVPR-23 & - & \checkmark & 22.29 & 46.72 &  88.82 & 93.15 & 62.75 \\
        \rowcolor{cyan!10}
        MaPLe + Ours & ViT/CLIP & - & - & \checkmark & 24.86 & 53.11 & 92.50 & 94.85 & 66.33 \\
        \textcolor{teal}{\small		\textbf{$\Delta$}} & -	& - & - & - &	\textcolor{teal}{\small	\textbf{+2.57}} & 	\textcolor{teal}{\small	\textbf{+6.39}} & 		\textcolor{teal}{\small		\textbf{+3.68}} & 		\textcolor{teal}{\small		\textbf{+1.70}} & 		\textcolor{teal}{\small		\textbf{+3.58}} \\  \hline
        LoRA-CLIP  & ViT/CLIP &  CVPR-24 &  - & \checkmark & 24.44 & 50.68 & 92.63 & 96.20 & 65.99 \\ 
        \rowcolor{cyan!10}
          LoRA-CLIP + Ours & ViT/CLIP & - &  - & \checkmark & 25.79 & \textbf{55.95} & \textbf{93.43} & \textbf{96.88} & \textbf{68.01} \\
        \textcolor{teal}{\small		\textbf{$\Delta$}} & -	& - & - & - &	\textcolor{teal}{\small	\textbf{+1.35}} & 	\textcolor{teal}{\small	\textbf{+5.27}} & 		\textcolor{teal}{\small		\textbf{+0.80}} & 		\textcolor{teal}{\small		\textbf{+0.48}} & 		\textcolor{teal}{\small		\textbf{+2.02}} \\
        \bottomrule
      \end{tabular}
    }
\end{table*}

\printcredits

\section*{Declaration of competing interest}
I confirm that no authors of this manuscript have any competing financial or non-financial interests, currently or previously, including serving in an editorial capacity for the journal we are submitting to.

\section*{Acknowledgments}
This work is supported by the National Natural Science Foundation of China under grant 62206102; the National Key Research and Development Program of China under grant 2024YFC3307900; the National Natural Science Foundation of China under grants 62436003, 62376103 and 62302184; the Key Research and Development Program of Hubei under grant 2026BEB042;  the Major Science and Technology Project of Hubei Province under grants 2025BAB011 and 2024BAA008; the Hubei Science and Technology Talent Service Project under grant 2024DJC078; the CCF-Tencent Rhino-Bird Open Research Fund (Grant No. RAGR20260108); and Ant Group through the CCF-Ant Research Fund. The computation is completed in the HPC Platform of Huazhong University of Science and Technology.



\printbibliography[title={References}]




\clearpage

\bio{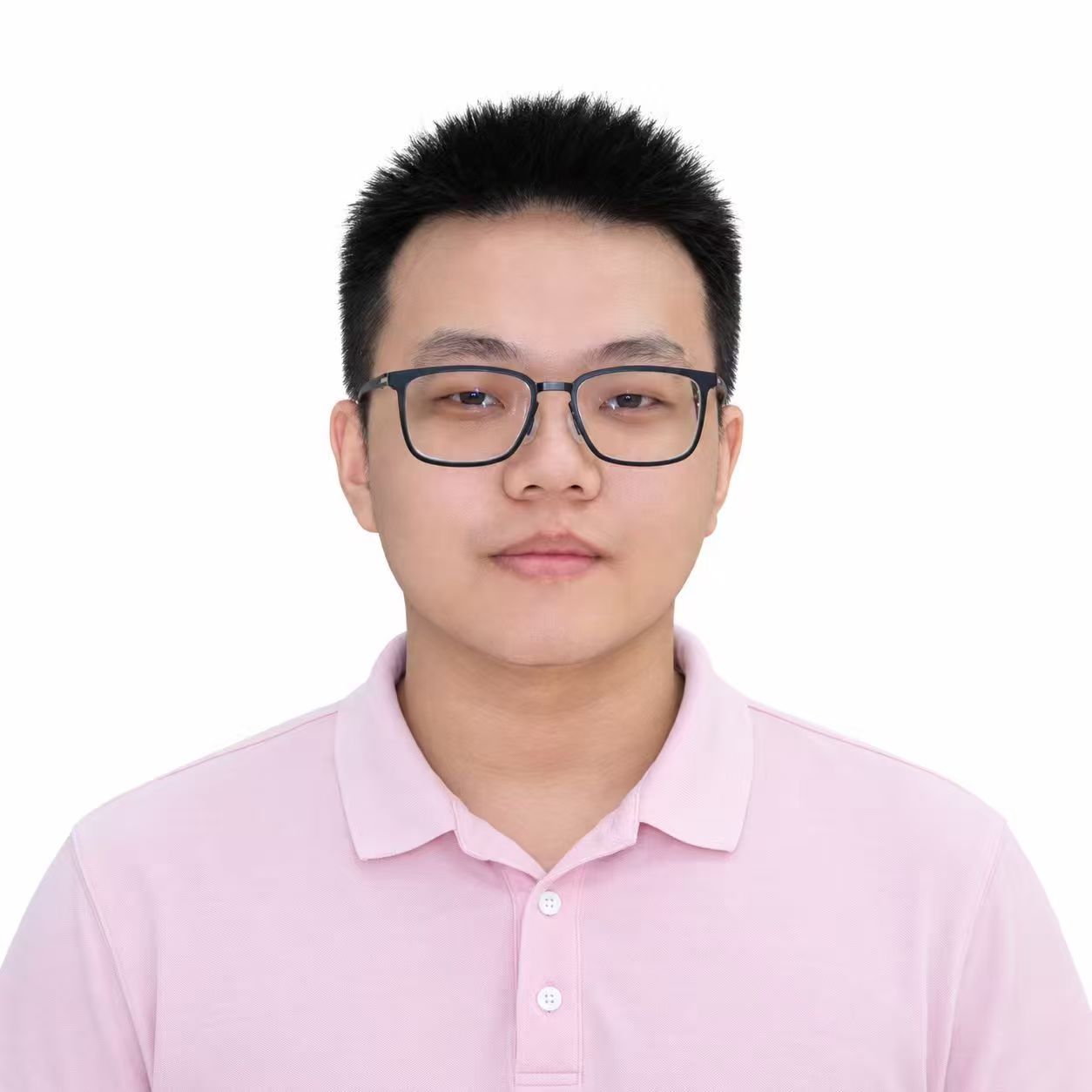}
\textbf{Yaze Zhao} received the B.S. degree from the School of Computer Science and Technology, Huazhong University of Science and Technology, Wuhan, China. He is also 
pursuing 
the master’s degree at the School of Computer Science and Technology, Huazhong University of Science and Technology. His primary research interests include few-shot learning, domain adaptation, and multimodal models.
\endbio

\bio{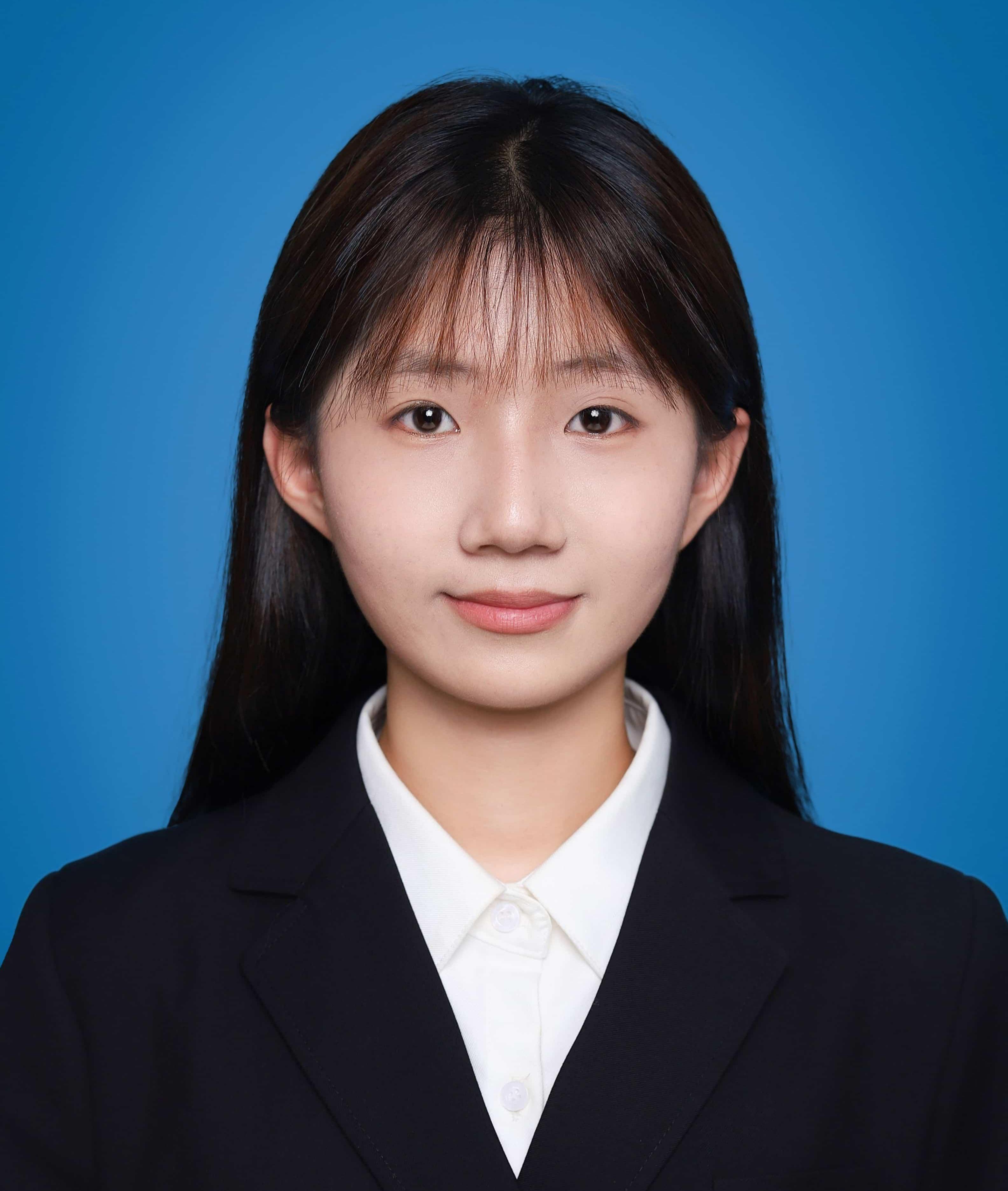}
\textbf{Yicong Liu} received the  B.S. degree from the College of Computer Science and Electronic Engineering, Hunan University. She received the M.S. degree from the School of Computer Science and Technology, Huazhong University of Science and Technology. Her primary research interests include few-shot learning and frequency domain analysis.
\endbio

\bio{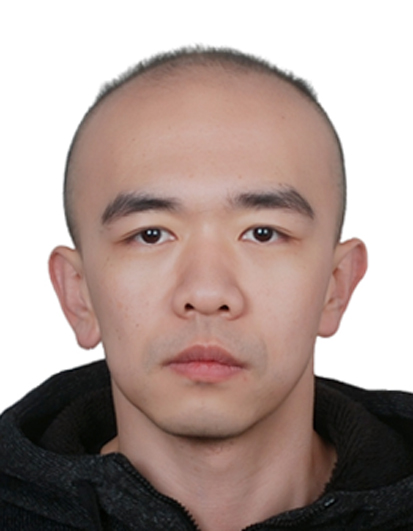}
\textbf{Yixiong Zou} received the B.S. degree from Nankai University, and received the Ph.D. degree from the School of Electrical Engineering and Computer Science, Peking University, Beijing, China. He was a visiting scholar at Carnegie Mellon University. He has published more than 40 journal and conference papers. He is currently a Lecturer with the School of Computer Science and Technology, Huazhong University of Science and Technology. His research interests include multimodal large language models, few-shot learning, open-world learning, and computer vision.
\endbio

\bio{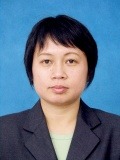}
\textbf{Yuhua Li} received the Ph.D. degree in computer application technology from Huazhong University of Science and Technology, Wuhan, China, in 2006. She is currently a Professor in the School of Computer Science and Technology, Huazhong University of Science and Technology. She was a visiting scholar at the University of California, Santa Barbara. She has published more than 60 journal and conference papers (NeurIPS, TKDE, SIGIR, WWW, ICDM, IJCAI). She is also a senior member of the China Computer Federation (CCF). Her research interests include data mining, social networks, machine learning, and big data.
\endbio

\bio{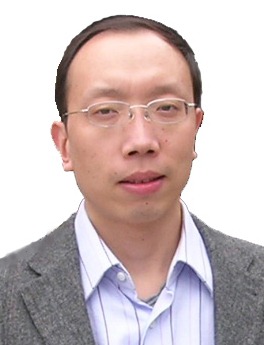}
\textbf{Ruixuan Li} received the B.S., M.S., and Ph.D. degrees in computer science from the Huazhong University of Science and Technology in 1997, 2000, and 2004, respectively. From 2009 to 2010, he was a Visiting Researcher with the Department of Electrical and Computer Engineering, University of Toronto. He is currently a Professor with the School of Computer Science and Technology, Huazhong University of Science and Technology. His research interests include cloud and edge computing, big data management, and distributed system security. He has published more than 500 journal and conference papers (NeurIPS, KDD, ICDM, IJCAI). He is also a member of ACM.
\endbio

\end{document}